\documentclass{article}
\usepackage{ijcai23}

\usepackage{times}
\usepackage{soul}
\usepackage{url}
\usepackage[hidelinks]{hyperref}
\usepackage[utf8]{inputenc}
\usepackage[small]{caption}
\usepackage{graphicx}
\usepackage{amsmath}
\usepackage{amsthm}
\usepackage{booktabs}
\usepackage{algorithm}
\usepackage{algorithmic}
\usepackage[switch]{lineno}
\usepackage{graphics}
\usepackage{multicol}
\usepackage{multirow}
\usepackage{xcolor}
\usepackage{amsmath}
\usepackage{amssymb}
\usepackage{amsthm}
\usepackage{enumitem}
\usepackage{bm}
\usepackage{booktabs}
\usepackage{units}
\definecolor{orange}{RGB}{255, 127, 0}

\theoremstyle{definition}
\newcommand{\modelname}{\texttt{HyperFed}}
\usepackage{subfigure}
\title{HyperFed: Hyperbolic Prototypes Exploration with Consistent Aggregation for Non-IID Data in Federated Learning}

\author{
Xinting Liao$^1$
\and
Weiming Liu$^1$\and
Chaochao Chen$^{1}$\thanks{Chaochao Chen is the corresponding author.}\and
Pengyang Zhou$^1$\and
Huabin Zhu$^1$\and\\
Yanchao Tan$^2$\and
Jun Wang$^3$\And
Yue Qi$^3$
\affiliations
$^1$College of Computer Science and Technology, Zhejiang University, Hangzhou, China\\
$^2$College of Computer and Data Science, Fuzhou University, Fuzhou, China\\
$^3$OPPO Research Institute, Shenzhen, China\\
\emails
\{xintingliao, 21831010, zjuccc, zhoupy, zhb2000\}@zju.edu.cn,
yctan@fzu.edu.cn,
junwang.lu@gmail.com,
qiyue@oppo.com
}
\begin{document}

\maketitle

\begin{abstract}
    Federated learning (FL) collaboratively models user data in a decentralized way.
    However, in the real world, 
    non-identical and independent data distributions (non-IID) among clients hinder the performance of FL due to three issues, i.e., (1) the class statistics shifting, (2) the insufficient hierarchical information utilization, and (3) the inconsistency in aggregating clients.
    To address the above issues, we propose \modelname~which contains three main modules, i.e., hyperbolic prototype Tammes initialization (HPTI), hyperbolic prototype learning (HPL), and consistent aggregation (CA).
     Firstly, HPTI in the server constructs uniformly distributed and fixed class prototypes, and shares them with clients to match class statistics, further guiding consistent feature representation for local clients.
    Secondly, HPL in each client captures the hierarchical information in local data with the supervision of shared class prototypes in the hyperbolic model space.
    % 
    % For handling issue 2, the hyperbolic prototype learning module in each client model local data in the hyperbolic model space with the supervision of fixed and shared class prototypes.
    % 
    Additionally, CA in the server mitigates the impact of the inconsistent deviations from clients to server. 
    Extensive studies of four datasets prove that \modelname~is effective in enhancing the  performance of FL under the non-IID setting.

\end{abstract}

\section{Introduction}
% background for FL, and why PFL
% Out of the privacy-leakage concerns,\xenia{del not main topic} 
Federated Learning (FL) trains a global model by collaboratively modeling decentralized data in local clients~\cite{mcmahan2017communication}.
Disappointingly, FL 
comes into a performance bottleneck in many real-world applications, where clients contain data with 
non-identical and independent distributions (non-IID) ~\cite{li2019convergence,zhao2018federated}. 
Taking a hand-written recognition system as an example, 
different people have their personalized writing styles, making hand-written characters and letters differ in shape, size, and so on.

\begin{figure}[h]
\centering
\includegraphics[width=0.9\linewidth]{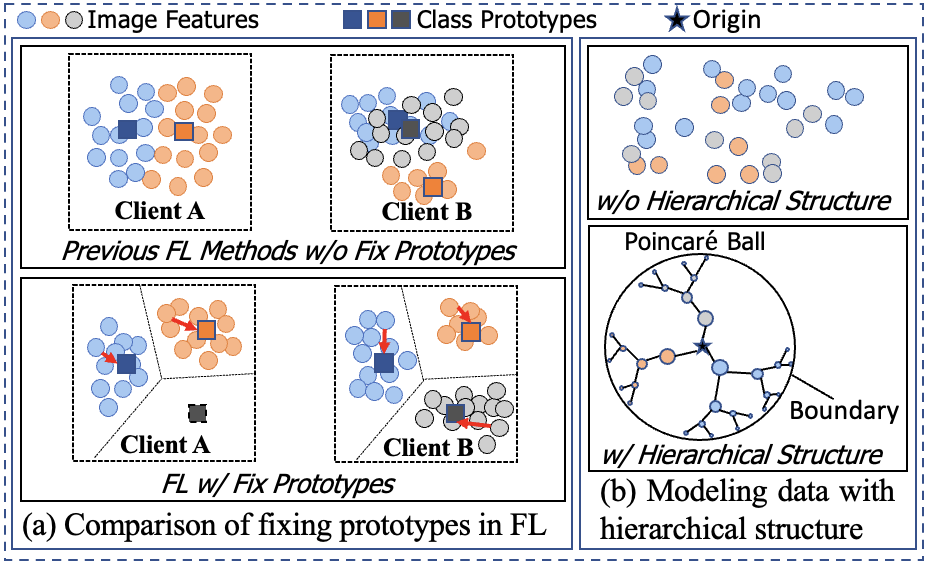} % Reduce the figure size so that it is slightly narrower than the column. Don't use precise values for figure width. This setup will avoid overfull boxes.
\caption{A motivation of fixed class prototypes in hyperbolic space.}
\label{fig:motivation}
\end{figure}
% \lwm{
% 1. Fig1, prototype shift; scattered; non-iid setting lead to ... SphereFed ...  }
% \lwm{
% 1. Fig1, hyperbolic space ... }

Existing FL work with non-IID data either improves the performance of the general global model or enhances the personalized local model.
First, to obtain better global performance, a number of work tries to modify the local objectives by adding regularization, so as to make them consistent with generic global performance, e.g., FedProx~\cite{li2020federated}.
Second, to enhance the performance of the local models, a series of studies encourage training a personalized model for individual clients with meta-learning~\cite{fallah2020personalized}, transfer learning\cite{luo2022disentangled}, and so on.
Recently, FedBABU~\cite{oh2021fedbabu} and Fed-RoD~\cite{chen2021bridging} find it possible to enhance global and local models simultaneously, which decouple model in FL into two parts, i.e., one for global generalization and the other for personalization.
% 
% Though decomposition brings blessings with better performance, it is non-trivial to constructing an ideal personalized predictor that is scalable and determinant to inference.

Nevertheless, most existing methods overlook three issues in FL with non-IID data.
Firstly, \textit{ class statistics shifting \textbf{(Issue 1)}} happens in FL with non-IID data.
Clients have different class statistics information of local data distributions, i.e., the class prototypes, which will shift and bring trivial solutions in FL without fixing.
As Fig.~\ref{fig:motivation}(a) shows, ignoring fixing class prototypes brings two inevitable limits.
Client A fails to recognize the missing class, i.e., data corresponding to some class is missing, while client B causes the class overlapping, i.e., gathering the prototypes of blue and gray classes too tight to discriminate their image features.
Though FedBABU~\cite{oh2021fedbabu} and SphereFed~\cite{dong2022spherefed} contribute to addressing this issue, they suffer from the \textit{dimension dilemma} problem, i.e., they either lack scalability in low-dimensional space, or generate sparse representation in high-dimensional space. 
Secondly, current work only captures the semantic information of non-IID data, having \textit{the insufficient hierarchical information utilization} issue \textbf{(Issue 2)}.
As depicted in Fig.~\ref{fig:motivation}(b), it is hard to gather the data samples of the same class together without hierarchical information.
Hierarchical information can be helpful to group data samples and generate fine-grained representations, which can further bring prediction gains.
To take this advantage, hyperbolic models are used for continuously capturing the hierarchical structure of data in the low-dimensional space~\cite{liu2020hyperbolic,linial1995geometry}, whose effectiveness is proved in computer vision~\cite{khrulkov2020hyperbolic}, recommender systems~\cite{tan2022enhancing}, and natural language processing~\cite{chen2022fully}. 
% 
% As depicted in Fig.~\ref{fig:motivation}(a), both Euclidean tree and hyperbolic model, e.g., Poincar\'e ball, can capture the partial order structure inherent in data, which tights the data samples of the same class together for determined prediction.
% 
% For example, it is vital for prediction with the structural information, since the partial-order relationship pulls the pairs of the same class closer than the pairs with different classes. \ccc{}
% 
% Compared with modeling data in Euclidean tree, hyperbolic model is more effective in continuously capturing the hierarchical structure of data in the low-dimensional space~\cite{liu2020hyperbolic,linial1995geometry}.
% 
However, how to integrate existing FL methods with the hierarchical information in hyperbolic space remains unknown.
Lastly, \textit{the inconsistency in aggregating clients \textbf{(Issue 3)}} deteriorates the performance of current FL methods as well. 
In practice, clients usually have statistically heterogeneous data.
% 
% It is practical to involve some clients with statistically heterogeneous data in real-world FL applications. \ccc{In practice, clients usually have statistically heterogeneous data. } 
% 
Existing aggregation methods, e.g., weighted average with data amounts, result in the aggregated global model deviating from the optimal global model. 

In this work, we propose \modelname~which contains three modules to address the above issues, respectively. 
% , which explores hyperbolic prototypes and consistency updating in FL. 
%
% Specifically, \modelname consists of three modules, i.e., \textbf{hyperbolic prototype Tammes initialization (HPTI)}, \textbf{hyperbolic prototype learning (HPL)}, and \textbf{consistent aggregation (CA)} to address three issues in non-IID setting.
% (sol2)
To solve \textbf{\textit{Issue 1}} and avoid dimension dilemma, \modelname~contains \textbf{hyperbolic prototype Tammes initialization (HPTI)} module in server.
The server first uses Tammes prototype initialization (TPI) to construct uniformly distributed class prototypes for the whole class set in hyperbolic space. 
Then the server fixes the position of class prototypes, and shares class prototypes to initialize the hyperbolic predictors of clients.
In this way, \modelname~not only guides the consistent and separated criteria, but also introduces the statistics of missing class, both of which encourage discriminative and fine-grained feature representation for non-IID data.
%(sol1) 
To avoid \textbf{\textit{Issue 2}}, 
% (advantage2) 
with the supervision of hyperbolic prototypes, \textbf{hyperbolic prototype learning (HPL)} module in each client pulls the data sample close to its ground truth class prototype, and pushes other data samples away.
Thus \modelname~enjoys the benefits of predicting with hierarchical information.
% (sol3) 
% To minimize the performance deterioration caused by 
To tackle 
\textbf{\textit{Issue 3}}, \modelname~has a \textbf{consistent aggregation (CA)} module that resolves the inconsistent deviations from clients to server by solving a multi-objective optimization with Pareto constraints.
% (advantage3) 
Hence \modelname~obtains consistent updating direction among clients, without the need of cumbersome grid search for aggregating clients.

In summary, we are the first, as far as we know, to explore hyperbolic prototypes in FL with non-IID Data. 
We contribute in:
(1)
We adopt uniformly distributed and fixed class prototypes in hyperbolic space to alleviate the 
impact of statistics shifting. 
(2)
We 
sufficiently leverage hyperbolic representation space to capture hierarchical information for FL.
(3) We optimize the aggregation of different client model to a Pareto stationary point, minimizing the impact of inconsistent clients deviations.
(4) Extensive experiments on four benchmark datasets prove the effectiveness of \modelname.

\section{Related Work}

\subsection{Federated Learning for Non-IID Data}
In terms of the goal of optimization, there are mainly three categories of common FL work that tackles non-IID data:
(1) Global performance, which modifies the local objectives with a regularization term to obtain a
well-performed global model~\cite{li2020federated}.
FedProx~\cite{li2020federated} proposes an additional proximal term to local objective, which penalizes the updated local model that is far away from the global model.
FedDYN~\cite{acar2020federated} and MOON~\cite{li2021model} regularize the model change with both historical global and local models simultaneously.
(2) Local performance, which trains a personalized model for individual clients to enhance the performance of local models~\cite{t2020personalized};
and (3) Global and local performance, which empirically decomposes the network in FL into the body for universal representation, and the head for personalized classification. 
 Fed-RoD~\cite{chen2021bridging} consists of two classifiers to maintain the local and global performance, respectively.
 However, no above work takes action to avoid class shifting.
 Few work fixes the classifier to fill this gap, e.g., FedBABU~\cite{oh2021fedbabu} and SphereFed~\cite{dong2022spherefed}.
FedBABU randomly initializes and fixes the classifier during training FL, which cannot guarantee the separation of different classes is distinguishable enough.
SphereFed considers fixing the classifier in hyperspherical space.
But SphereFed either lacks scalability in low-dimensional space, or generates sparse representation in high-dimensional space.
VFGNN~\cite{CCIjcai22} utilizes graph sturcture in vertical FL rather than horizontal FL.
Moreover, few existing work considers utilizing hierarchical structure and consistent aggregation, which degrades FL with non-IID data.

\subsection{Hyperbolic Representation Learning}
Hyperbolic geometry is a non-Euclidean geometry, which can be constructed by various isomorphic models, e.g., Poincar\'e model~\cite{nickel2017poincare}. 
Hyperbolic modeling has been leveraged in various deep networks, such as fully-connected layers~\cite{shimizu2020hyperbolic}, convolutional layers~\cite{shimizu2020hyperbolic}, recurrent layers~\cite{ganea2018hyperbolic}, classification layers~\cite{cho2019large,weber2020robust}, graph neural networks~\cite{liu2019hyperbolic,tan2022towards} and Transformer~\cite{ermolov2022hyperbolic}.
However, the existing work overlooks taking the advantage of hyperbolic learning in FL.
\cite{shen2021spherical,mettes2019hyperspherical,ghadimi2021hyperbolic}
treat additional prior using orthogonal basis or the prior knowledge 
 embeddings as hyperbolic prototypes, which are positioned as distant as possible from the origin, to avoid frequently updating in prototype learning.
These motivate us to utilize hyperbolic prototypes
% take the hyperbolic prototypes as class prototypes 
in \modelname.
On the contrary, we contract the class prototypes away from the bound of the Poincar\'e ball model, 
% for the purpose of 
aiming to obtain more general class semantic information.

\section{Preliminary: Poincar\'e Ball Model}
% 用自己的话表达 Hyperbolic Image Segmentation
% ref Hyperbolic visual embedding learning for zero-shot recognition
Poincar\'e ball model is one of the common models in hyperbolic space, which is a type of Riemannian manifold $\mathcal{M}$ with constant negative curvature.
In this work, $||\boldsymbol{x}||_2 = \sqrt{\Sigma_{i=1}^n x_i^2}$ is a Euclidean norm.
The Poincar\'e is an open ball model in $n$-dimensional hyperbolic space defined as $(\mathbb{P}^n, g^{\mathbb{P}})$, where $\mathbb{P}^n = \{\boldsymbol{x}\in \mathbb{R}^n: ||\boldsymbol{x}||_2 < 1\}$ 
and $g^{\mathbb{P}}$ is the Riemannian metric of a Poincar\'e ball.
$g^{\mathbb{P}}$ is conformal to the metric of Euclidean space $g^\mathbb{E}$, i.e.,
$
g^{\mathbb{P}}_{\boldsymbol{x}} = \lambda_{\boldsymbol{x}}^2 g^\mathbb{E},
$
where $\lambda_{\boldsymbol{x}}=\frac{2}{1-||\boldsymbol{x}||_2^2}$ is the conformal factor.
% \xenia{and $g^\mathbb{E}$ is the metric of Euclidean space}.
%
Given two points in the Poincar\'e ball model, i.e., $\boldsymbol{x}_1, \boldsymbol{x}_2\in \mathbb{P}^n$, the M\"obious addition is defined as:
$$\boldsymbol{x}_1\oplus \boldsymbol{x}_2 = \frac{(1+2\langle\boldsymbol{x}_1, \boldsymbol{x}_2\rangle + ||\boldsymbol{x}_2||_2^2)\boldsymbol{x}_1 +(1-||\boldsymbol{x}_1||_2)\boldsymbol{x}_2^2}{1+2\langle\boldsymbol{x}_1, \boldsymbol{x}_2\rangle + ||\boldsymbol{x}_1||_2^2||\boldsymbol{x}_2||_2^2 }$$
We can define the geodesic, i.e., the shortest distance between these two points in the Poincar\'e ball as below:
\begin{equation}\label{eq:geo_dis}
    d(\boldsymbol{x}_1, \boldsymbol{x}_2) = \operatorname{arcosh}(1 + \frac{2||\boldsymbol{x}_1-\boldsymbol{x}_2||_2^2}{(1 -  ||\boldsymbol{x}_1||_2^2)(1 - ||\boldsymbol{x}_2||_2^2)}).
\end{equation}
For a point $\boldsymbol{x}$ in a manifold $\mathcal{M}$, the tangent space $T_{\boldsymbol{x}} \mathcal{M}$ is a vector space comprising all directions that are tangent to $\mathcal{M}$ at $\boldsymbol{x}$.
% \xenia{For a point $x$ in a manifold $\mathcal{M}$, one can define the tangent space $T_x \mathcal{M}$ of $\mathcal{M}$ at $x$ as a vector space that contains all possible directions in which one can tangentially pass through. An inner product can be defined on $T_x \mathcal{M}$. }
% 
Exponential map $\exp_{\boldsymbol{x}}: T_{\boldsymbol{x}} \mathcal{M} \rightarrow \mathcal{M}$ is a transformation that 
projects any point $\boldsymbol{u}$ from the Euclidean tangent space to the Poincar\'e ball referred by point $\boldsymbol{x}$, defined as:
\begin{equation}
\exp _{\boldsymbol{x}}(\boldsymbol{u})=\boldsymbol{x} \oplus\left(\tanh \left(\frac{\lambda_{\boldsymbol{x}}\|\boldsymbol{u}\|_2}{2}\right) \frac{\boldsymbol{u}}{\|\boldsymbol{u}\|_2}\right).
\end{equation}
The reverse of $\exp_{\boldsymbol{x}}$ is logarithmic map, i.e., $\log _{\boldsymbol{x}}: \mathcal{M} \rightarrow T_{\mathbf{x}} \mathcal{M}$, which projects hyperbolic vector back to Euclidean space. 
\section{Method}

\subsection{Problem Statement}
% ref Model-Contrastive Federated Learning
% ref https://www.aaai.org/AAAI21Papers/AAAI-5802.HuangY.pdf
% ref Learning to Generalize in Heterogeneous Federated Networks
In FL with non-IID setting, we assume there are $K$ clients, containing their own models and local datasets, and a central server with global model aggregated from clients.
Suppose a dataset $\mathcal{D}$ has $C$ classes indexed by $[C]$, where $[C]$ means the full set of labels in $\mathcal{D}$.
% , and each sample in $\mathcal{D}$ denotes as $(\boldsymbol{x}, y) \in \mathcal{X} \times [C]$.}
%
Each client $k$ has access to its private local dataset $\mathcal{D}_k=\{ \boldsymbol{x}_{k,i}, y_{k,i}\}_{i=1}^{N_k}$, containing $N_k$ instances sampled from distinct distributions.
Therefore, we get $\mathcal{D} =\cup_{k\in [K]} \mathcal{D}_k$, where the data distributions of different $\mathcal{D}_k$ are different.
The overall objective in FL with non-IID setting is defined as below: 
\begin{equation}
\small
{\operatorname{min}} \mathcal{L}(\boldsymbol{\theta}_{1}, \dots, \boldsymbol{\theta}_{K}; \boldsymbol{p})=\Sigma_{k=1}^K p_k \mathbb{E}_{(\boldsymbol{x}, {y})\sim D_k}[\mathcal{L}_k(\boldsymbol{\theta}_{k}; (\boldsymbol{x}, {y})],
\label{eq:pfl_problem}
\end{equation}
where $\mathcal{L}_k (\cdot)$ is the model loss at client $k$, and $p_k$ indicates its weight ratio for aggregating.
% makes all of the local Poincar\'e models coincide with each other

\begin{figure}[t]
\centering
\includegraphics[width=\linewidth]{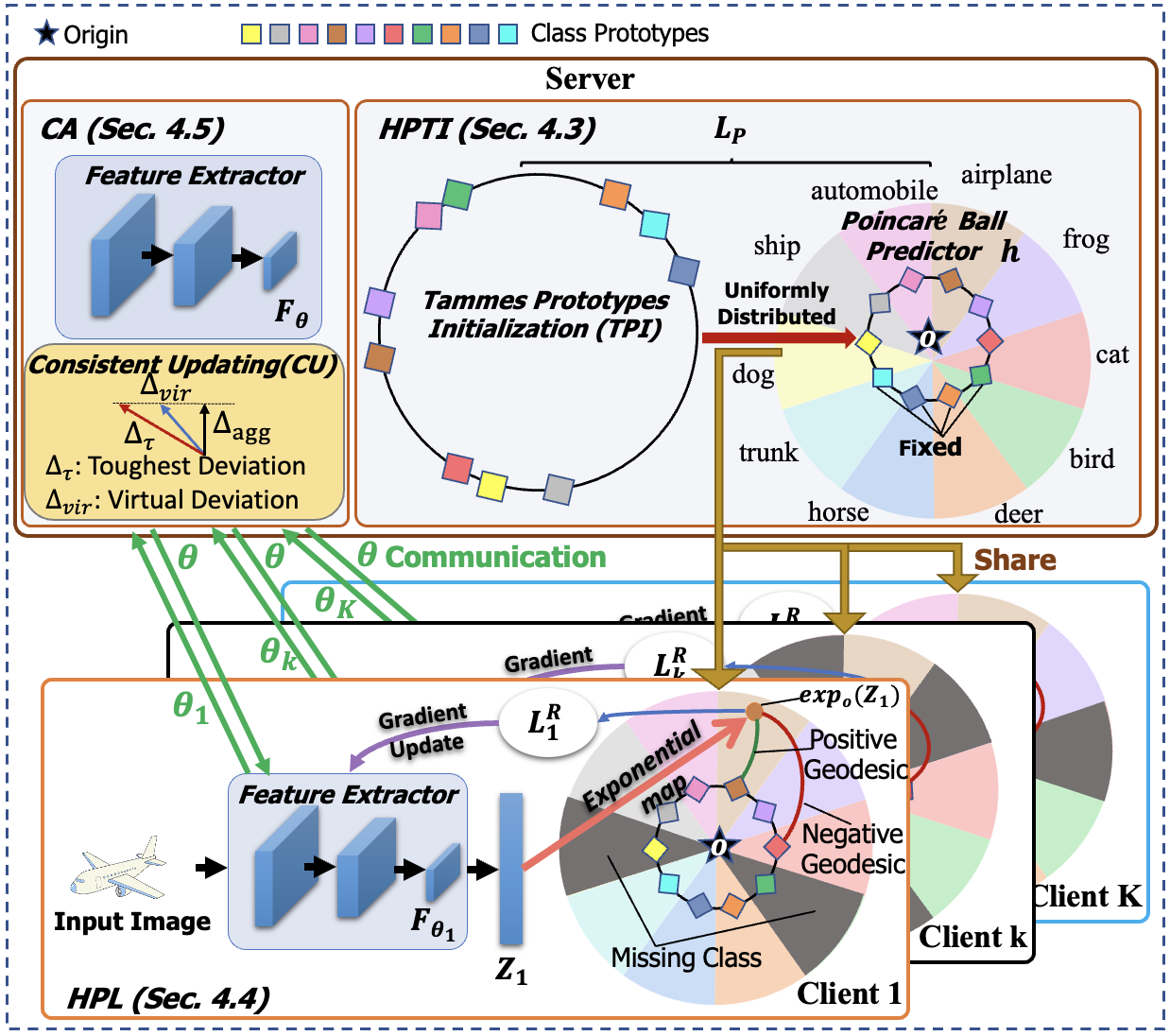} % Reduce the figure size so that it is slightly narrower than the column. Don't use precise values for figure width. This setup will avoid overfull boxes.
\caption{Framework of \modelname. 
We take the modeling of \modelname~with 2-dimensional Poincar\'e ball predictor on Cifar10 as an example. }
\label{fig:HyperbolicFed}
\end{figure}

\subsection{Framework Overview}
To explain how \modelname~solve the problem in FL with non-IID data, i.e., Eq.~\eqref{eq:pfl_problem}, we introduce the framework overview of \modelname.
In Fig.~\ref{fig:HyperbolicFed}, there are a server and K clients.
Each client or server, similarly consists of a feature extractor, an exponential map, and a Poincar\'e ball predictor. 
The feature extractor $\mathcal{F}(\cdot): \mathcal{X}\rightarrow \mathbb{R}^d$ maps an input data $\boldsymbol{x}$ into a $n$-dimensional vector $\boldsymbol{z}=\mathcal{F}(\boldsymbol{x})$ as feature representation.
Then we get $\operatorname{exp}_{\boldsymbol{o}} (\boldsymbol{z})$ by
leveraging exponential map on the feature representation $\boldsymbol{z}$ to the Poincar\'e ball space referred by the origin $\boldsymbol{o}$. 
Finally, the Poincar\'e ball predictor $h(\cdot):\mathbb{R}^d \rightarrow \mathcal{Y}$ decides class label for input data based on the representation $\operatorname{exp}_{\boldsymbol{o}} (\boldsymbol{z})$ in Poincar\'e ball. 
All Poincar\'e ball predictors of server and clients are fixed and shared. % in FL modeling.

As Fig.~\ref{fig:HyperbolicFed} shows, there are mainly three steps in  \modelname.
(1) The server in \modelname~leverages hyperbolic prototype Tammes
initialization (HPTI) module to construct a full set of uniformly-distributed class prototypes 
for the Poincar\'e ball predictor by Tammes prototype initialization (TPI), contracts class prototypes close to origin,
and shares Poincar\'e ball predictor with fixed class prototypes to all of the clients.
(2) Each client models local data distribution independently with the hyperbolic prototype learning (HPL) module, then sends the parameters of the local model to the server for aggregation.
(3) Consistent aggregation (CA) module in the server updates the global model parameters using consistent updating (CU) to mitigate the inconsistent deviations from clients to server. After that, the server sends the new global model parameters back to clients.
This communication between server and clients, i.e., steps 2-3, iterates until the performance converges.

\subsection{Hyperbolic Prototype Tammes Initialization}
% \subsection{Hyperbolic Prototype Modelling}
\paragraph{Motivation.} In this section, we devise HPTI module in server to resolve two limits brought from the class statistics shifting, i.e., missing class and class overlapping, as described in Fig.~\ref{fig:motivation}(a).
To bypass the dilemma of choosing a dimension, HPTI explores the class statistics in the hyperbolic space, which is scalable and effective in modeling data with low-dimensional space. 
Firstly, HPTI uses Tammes prototype
initialization (TPI) to construct uniformly distributed and distinguishable class prototypes for the entire class set. 
Then HPTI fixes the position of the class prototypes on the Poincar\'e ball predictor.
Lastly, HPTI sends the Poincar\'e ball predictor with fixed hyperbolic class prototypes, including the missing classes, to clients.
% , obtaining consistent criteria for local modeling in the next section. 
% 
We describe them in detail below.

% % 
HPTI first constructs uniformly distributed hyperbolic prototypes with TPI, which is available in data without prior semantic information and efficient in computation.
Most work relies on prior semantic knowledge about classes to discover the positions and representations of class prototypes.
However, not all datasets contain prior semantic knowledge.
Motivated by \cite{ghadimi2021hyperbolic}, we randomly sample points on the boundary of Poincar\'e ball for assigning class prototypes, and optimize these points to be uniformly distributed in a ball-shaped space.
In this way, we incorporate the prior with large margin separation for Poincar\'e ball predictor.
To be specific, searching for uniformly distributed class prototypes can be formulated as a Tammes problem~\cite{tammes1930origin}, i.e., arranging a set of points on a unit sphere that maximizes the minimum distance between any two points. 
TPI optimizes this Tammes problem to obtain the class prototypes for all $C$ classes in a dataset, i.e., $\boldsymbol{W}^* \in \mathbb{R}^{C\times n}$ with $n$ denoting the dimension of prototype:
\begin{equation}
\small
\begin{gathered}
\label{eq:tammes}
\boldsymbol{W}^*=\underset{\boldsymbol{W} \in \mathbb{P}^n}{\operatorname{argmin}}\left(\max _{(i, j, i \neq j) \in[C]} \boldsymbol{w}_i {\boldsymbol{w}_j}^{\top}\right), \\
\text { s.t. } \forall i \in [C] \quad\left\|\boldsymbol{w}_i\right\|_2=1,
\end{gathered}
\end{equation}
where $\boldsymbol{w}_i$ ($\boldsymbol{w}_j$) is the $i-$th ($j-$th) row of $\boldsymbol{W}$ representing as the $i-$th ($j-$th) class prototype.
We choose cosine similarity to measure this distance,
because Poincar\'e ball model space is conformal to the Euclidean space~\cite{ganea2018hyperbolic}.

Optimizing Eq.~\eqref{eq:tammes} requires computing pairwise similarity of class prototypes iteratively, which is inefficient.
To mitigate it, we utilize the similarity of Poinecar\'e ball and hyper-sphere to follow~\cite{mettes2019hyperspherical},  and minimize the largest cosine similarity for each prototype in the form of matrix, thus accelerating the optimization:
\begin{equation}
\small
\begin{gathered}
\mathbf{L}_{P}=\frac{1}{C} \Sigma_{i=1}^C \max _{j \in [C]} \mathbf{M}_{i j}, \mathbf{M}=\boldsymbol{W} \boldsymbol{W}^T-2 \mathbf{I}, \\
\text{ s.t. } \forall i \in [C] \quad \left\|\boldsymbol{w}_i\right\|_2=1.
\end{gathered}
\end{equation}

Next, we find the position to fix the hyperbolic class prototypes.
As Fig.~\ref{fig:motivation}(b) shows, in Poincar\'e ball model, the closer the distance from the referred origin to the node, the more general the semantic information of node represents~\cite{liu2020hyperbolic}.
But the uniformly distributed class prototypes are initially positioned on the boundary of the Poincar\'e ball, which is against the distribution of hierarchical structure in Poincar\'e ball.
In order to enjoy the benefits of uniformity and generality simultaneously, we contract the class prototypes along with the radius to the origin $\boldsymbol{o}$ by a slope degree $s$, i.e., $\boldsymbol{W}_{\mathbb{P}}= s 
 \boldsymbol{W}^* \in \mathbb{R}^{C\times n}$.
Lastly, HPTI~shares and fixes the Poincar\'e ball model to clients, which encourages local clients to model local data sufficiently with the supervision of consistent and separated hyperbolic prototypes.

% 
% Hence, each client can consistently optimize objective in Eq.~\eqref{eq:loss} and predict its local data samples according to the geodesic distance, i.e., Eq.~\eqref{eq:geo_dis}, between themselves and the shared class prototypes:
% \begin{equation}
% \label{eq:pred}
% y^{\star}=\arg \min _{\boldsymbol{w}_{y} \in \boldsymbol{W}_{\mathbb{P}}} d \left(\exp _{\boldsymbol{o}}(\mathcal{F}_{\boldsymbol{\theta}} (\mathbf{x})), \boldsymbol{w}_{y}\right).
% \end{equation} 
% 
 
\subsection{Hyperbolic Prototype Learning}
\paragraph{Motivation.}
In this part, we provide the details of HPL, which utilizes the hierarchical information inherent in data to obtain fine-grained and gathered data representations.
To utilize the hierarchical information of data, HPL uses Poincar\'e ball model for the benefits of continuous optimization and effective representation in low-dimensional space.
To start with, HPL extracts the feature for data samples and applies an exponential map referred by the origin shared with class prototypes. 
According to the supervision of shared class prototypes in hyperbolic space, HPL next represents the data features of the same class according to hyperbolic triplet loss.
% 

% \nosection{Personalized  
% Hyperbolic Representation}
% 【每个技术点都需要介绍清楚，它是什么，因为什么原因要用或者怎么起到了作用的】
% P1: 生成prototype ref busemann 
% 
In the following, we present how to model the hyperbolic representations of data samples for each client $k$ locally. 
Specifically, we expect to learn a projection of local data $\mathcal{D}_k$ to the local Poincar\'e ball model, i.e., $\mathbb{P}^n_k$, in which we compute the similarity between data samples and class prototypes for each client.
As introduced in Fig.~\ref{fig:HyperbolicFed}, we take a feature extractor $\mathcal{F}_{\boldsymbol{\theta}_k}(\cdot)$ in Euclidean space to obtain the feature representations of local data samples, i.e., $\boldsymbol{z}=\mathcal{F}_{\boldsymbol{\theta}_k}(\boldsymbol{x})$ for an instance pair $(\boldsymbol{x}, y)$ in $\mathcal{D}_k$.
Referred by origin $\boldsymbol{o}$, we apply an exponential map from tangent space $\mathcal{T}_{\boldsymbol{o}} \mathcal{M}$ to the Poincar\'e ball model $\mathbb{P}^n_k$ shared with class prototypes.
Hence, the representation of data samples in Poincar\'e ball model $\mathbb{P}^n_k$ can be:
\begin{equation}\label{eq:exp_map}
\exp_{\boldsymbol{0}}(\boldsymbol{z})=\tanh(||\boldsymbol{z}||_2)\frac{\boldsymbol{z}}{||\boldsymbol{z}||_2}.
\end{equation}

As mentioned ahead, we seek to construct the hierarchical structure between the feature representations of data samples and their corresponding class prototypes.
Triplet loss~\cite{movshovitz2017no,weimingNeurips21,weimingWWW23}
optimizes the distances among a set of triplets, denoted as \{anchor point, positive point, negative point\}, by creating a fixed margin, i.e., $m$, between the anchor-positive points difference and the anchor-negative points difference.
%
% 
% % 
% In detail, triplet loss takes a hinge function to create a fixed margin, i.e., $m$, between the anchor-positive points difference and the anchor-negative points difference:
% \begin{small}
%     \begin{equation}
% \mathcal{L}_{\text{triplet}}
% % (a,p_\text{pos},p_\text{neg})
% = \max(\operatorname{diff}(a,p_\text{pos})+m-\operatorname{diff}(a,p_\text{neg}),0 ). 
% \end{equation}
% \end{small}
% % 
Motivated by this, we choose each data sample representation in the Poincar\'e ball model as \textit{anchor point}, the ground truth class prototype as \textit{positive point}, and the remaining prototypes of the full class set as \textit{negative points}.
In this way, each client incorporates the prototypes of its missing class to feature representation, by randomly sampling negative points.
% \xenia{todel:In this way, the loss function for data sample $(\boldsymbol{x}, y) \in \mathcal{D}_k$ is free of the limitation of missing classes of data.} 
% 
We define hyperbolic triplet loss for client $k$ as below:
% \begin{align}
%     \mathcal{L} &= \max(d_{pos}-d_{neg}  + m, 0) \nonumber\\
%     d_{pos} &= d(\exp_{\boldsymbol{0}}(g_k(\boldsymbol{x}_i)), W_{k, y_i}) \nonumber\\
%     d_{neg} &= d(\exp_{\boldsymbol{0}}(g_k(\boldsymbol{x}_i)), W_{k, y_j}) \nonumber
% \end{align}
\begin{equation}
\small
    \label{eq:loss}
\mathbf{L}_{k}^{R}
% (\boldsymbol{z}, \boldsymbol{W}_{\mathbb{P}})
=\max(d(\exp_{\boldsymbol{0}}(\boldsymbol{z}), \boldsymbol{w}_y)-d(\exp_{\boldsymbol{0}}(\boldsymbol{z}), \boldsymbol{w}_{y^{\prime}})  + m, 0),
\end{equation}
where $\boldsymbol{z} = \mathcal{F}_{\boldsymbol{\theta}_k}(\boldsymbol{x})$, $\boldsymbol{w}_{y^{\prime}}$ is randomly sampled negative class prototype, $d(\cdot,\cdot)$ is the geodesic distance defined in Eq.~\eqref{eq:geo_dis}, and margin $m$ is a hyper-parameter.
We obtain fine-grained representation with sufficient hierarchical information, by simultaneously minimizing the positive geodesic, e.g., the green curve in Fig.~\ref{fig:HyperbolicFed}, and maximizing the negative geodesic, e.g., the red curve in Fig.~\ref{fig:HyperbolicFed}.
In this way, \modelname~utilizes the data hierarchical information to enhance the prediction.
% for FL with non-IID data.

\begin{algorithm}[h]
    \caption{Training procedure of \modelname}
    \label{alg:HyperFed}
    \textbf{Input}: Batch size $B$, communication rounds $T$, number of clients $K$, local steps $E$, dataset $\mathcal{D} =\cup_{k\in [K]} \mathcal{D}_k$\\
    % , where $\mathcal{D}_k=\{ \boldsymbol{x}_{k,i}, y_{k,i}\}_{i=1}^{N_k}$ \\
    \textbf{Output}: hyperbolic class prototypes $\boldsymbol{W}_{\mathbb{P}}$, model parameters, i.e., $\boldsymbol{\theta}^T \text{and} \{\boldsymbol{\theta}_k^T\}$
    \begin{algorithmic}[1]
        \STATE \textbf{Server executes():}
        \STATE Initialize 
 $\boldsymbol{\theta}^{0}$ with random distribution and $\boldsymbol{W}_{\mathbb{P}}$ by \textit{\textbf{HPTI}} 
        % \STATE Send model parameters to all participating clients
        \FOR{$t=0,1,...,T-1$}
            \FOR{$k=1,2,...,K$ \textbf{in parallel}} 
                \STATE Send $\{\boldsymbol{\theta}^t,\boldsymbol{W}_{\mathbb{P}}\}$ to client $k$ if $t=0$ else $\boldsymbol{\theta}^t$
                \STATE $\boldsymbol{\theta}_k^{t+1} \leftarrow$ \textit{\textbf{HPL:}} \textbf{Client executes}($k$, $\boldsymbol{\theta}^t$)
            \ENDFOR
        \STATE \textit{\textbf{CA:}} optimize Eq.~\eqref{eq:mgda} with CU and update parameters of $\boldsymbol{\theta}^{t+1}$ by Eq.~\eqref{eq:agg}
        \ENDFOR
        \STATE \textbf{return} $\boldsymbol{\theta}^T,\boldsymbol{W}_{\mathbb{P}}, \{\boldsymbol{\theta}_k^T\}$
        % \STATE
        \STATE \textit{\textbf{HPL:}} \textbf{Client executes}($k$, $\boldsymbol{\theta}^t$)\textbf{:}
        \STATE Assign global model to the local model $\boldsymbol{\theta}_k^t \leftarrow \boldsymbol{\theta}^t$
        \FOR{each local epoch $e= 1, 2,..., E$}
            \FOR{batch of samples $(\boldsymbol{x}_{k, 1:B}, \boldsymbol{y}_{k, 1:B}) \in \mathcal{D}_{k}$}
                \STATE Feature extraction $\boldsymbol{z}_{k, 1:B} \leftarrow \mathcal{F}_{\boldsymbol{\theta}_k^e} (\boldsymbol{x}_{k, 1:B})$
                \STATE Project $\boldsymbol{z}_{k, 1:B}$ to Poincar\'e ball by Eq.~\eqref{eq:exp_map}
                \STATE Compute loss $\mathbf{L}_k^{R}$ by Eq.~\eqref{eq:loss}
                % \STATE $d_{pos} \leftarrow d(\exp_{\boldsymbol{0}}(g_k(\boldsymbol{x}_i)), W_{k, y_i})$ 
                % \STATE $d_{neg} \leftarrow d(\exp_{\boldsymbol{0}}(g_k(\boldsymbol{x}_i)), W_{k, y_j}), y_j \ne y_i$  
                % \STATE $\mathcal{L} \leftarrow max(d_{pos} - d_{neg} + m, 0)$
                % \STATE $g_k^{t} \leftarrow \exp_{g_k^t}(-\eta_t\Delta_R \mathcal{L})$
                \STATE Update parameters of $\boldsymbol{\theta}_k^e$
                by RSGD
                % ~\cite{bonnabel2013stochastic} %Eq.~\eqref{eq:RSGD}
            \ENDFOR
        \ENDFOR
        \STATE \textbf{return} $\boldsymbol{\theta}_k^E$ to server
    \end{algorithmic}
\end{algorithm}

\subsection{Consistent Aggregation}
% \lwm{Adaptive Strategy}
% Lastly, the conflicting deviations of the extremely
% divergent clients (Issue 3) deteriorates the current FL meth-
% ods as well. It is inevitable to include some clients with ex-
% tremely statistically heterogeneous data distribution in the FL
% for real-world applications. The extremely divergent client
% will cause the heuristic aggregation, e.g., averaging weighted
% by the data amounts, to depart from the optimal global model,
% thus impacting the sub-stream training of clients.
% 
\paragraph{Motivation.}
Finally, we introduce CA which mitigates the inconsistent deviations from clients to server caused by the statistically heterogeneous data distributions.
In FL aggregation,
% \xenia{In FL aggregation,}
% 
CA first formulates the aggregation of local feature extractors in \modelname~ 
as a multi-objective optimization.
% , i.e., alleviating the inconsistent deviations from different clients to server. 
% 
Then CA applies consistent updating (CU) to pick the toughest client, i.e., the client with the most divergent deviation, and alleviate the inconsistency between the toughest client and the remaining clients. 
Lastly, CU iteratively optimizes this multi-objective optimization to yield a Pareto optimal solution and obtain the 
weight ratio of different client models.
% model parameter weights of different clients.
% 

We formulate the aggregation as a multi-objective optimization in the following. 
Specifically, we first compute the different deviations from clients to server as multiple objectives, then the goal of alleviating the inconsistency of these deviations can be achieved by multiple-objective optimization.
We obtain the combination of local parameters in server: 
\begin{equation}
\label{eq:agg}
\boldsymbol{\theta}^{t+1} = \boldsymbol{\theta}^{t}+\Sigma_{k=1}^K p_k\left(\boldsymbol{\theta}_k^{t+1}-\boldsymbol{\theta}^{t}\right),
\end{equation}
where $\boldsymbol{\theta}^{t}$ is the global model and $\boldsymbol{\theta}_k^{t}$ is the local model of the client $k$ at $t-$th communication.
Next, we denote global and client deviations, i.e., $\Delta_{\boldsymbol{\theta}}^{t+1}=\boldsymbol{\theta}^{t+1} - \boldsymbol{\theta}^{t}$ and $\Delta_{\boldsymbol{\theta}_{k}}^{t+1}=\boldsymbol{\theta}_k^{t+1}-\boldsymbol{\theta}^{t}$, respectively, and rewrite Eq.~\eqref{eq:agg} as:
% to obtain Eq.~\eqref{eq:agg} in the form of deviations:
\begin{equation}
\label{eq:agg_grad}
    \Delta_{\boldsymbol{\theta}}^{t+1} = \Sigma_{k=1}^K p_k \Delta_{\boldsymbol{\theta}_{k}}^{t+1}.
\end{equation}
% 
% The goal of consistent optimization is to alleviate the inconsistency of updating deviations from the different local models to the global model.
% % 
% Taking the different deviations from clients to server as multiple objectives, the goal of alleviating the inconsistency of these devitations can be achieved by multiple-objective optimization.
%
Then CA solves this multiple-objective optimization to Pareto stationary point, i.e., minimizing the minimum possible convex combination of inconsistent deviations:
\begin{small}
    \begin{equation}
\min \frac{1}{2}\left\| \Sigma_{k=1}^K p_k \Delta_{\boldsymbol{\theta}_{k}}^{t+1}\right\|^2_2 \text {, s.t. }  \Sigma_{k=1}^K p_k=1 \text {, and } \forall k, p_k \geq 0.
\label{eq:mgda}
\end{equation}
\end{small}

Next, we introduce CU~which derives from Multiple Gradient Descent Algorithm (MGDA)~\cite{desideri2012multiple,sener2018multi} to solve this optimization.
The optimization problem defined in Eq.~\eqref{eq:mgda} is equivalent to finding a minimum-norm point in the convex hull of the set of input points, i.e., a convex quadratic problem with linear constraints.
CU iteratively optimizes Eq.~\eqref{eq:mgda} by linear search, which can be solved analytically~\cite{jaggi2013revisiting}.
In detail, we find the toughest client, treat the combination of the remaining clients as a virtual client, and analyze the solution according to the directions of the toughest client and the virtual client. 
Firstly,
we initialize $\boldsymbol{p}_0$ with the weight of data samples, i.e., $p_k= \nicefrac{N_k}{\Sigma_{k=1}^{K}N_k} $, and 
precompute the consistency of deviations $\boldsymbol{V}, s.t. \boldsymbol{V}_{k,k^{\prime}}= {\Delta_{\boldsymbol{\theta}_{k}}}^{\top}{\Delta_{\boldsymbol{\theta}_{k^{\prime}}}}$. 
Then we find the toughest client by $\tau = \operatorname{arg}\operatorname{min}_{k^{\prime}} \Sigma_{k=1}^{K} p_k\boldsymbol{V}_{k^{\prime},k}$ with deviation $\Delta_{\tau}$, and remain the combination of others with historical weights to be a virtual client, i.e., $\Delta_{\text{vir}} = \Sigma_{k=1, k\neq \tau }^{K} p_k \Delta_{\boldsymbol{\theta}_{k}}$.
Thus we simplify Eq.~\eqref{eq:mgda}, i.e., $\operatorname{min}_{p_{\tau}\in [0,1]} \frac{1}{2} \|p_{\tau}\Delta_{{\tau}}+(1-p_{\tau})\Delta_{{\text{vir}}}\|^2_2$.
According to the directions of $\Delta_{\tau}$ and $\Delta_{\text{vir}}$, we can obtain the analytical solution for $p_{\tau}$:
\begin{equation}
    p_{\tau}= \operatorname{CU}^{+}\left[\nicefrac{{(\Delta_{\text{vir}}-\Delta_{\tau})}^{\top} \Delta_{\text{vir}}}{\|\Delta_{\tau}-\Delta_{\text{vir}}\|^2_2}\right],
    \label{eq:p_two_clients}
\end{equation}
where $\operatorname{CU}^{+}[\cdot] = \operatorname{max}(\operatorname{min} (\cdot, 1),0)$.
In Appendix C, we present the analysis of optimization for obtaining $\operatorname{CU}^{+}[\cdot]$  based on the computational geometry~\cite{sekitani1993recursive} of minimum-norm in the convex hull. 
Given $p_{\tau}$, we update the weight ratio $\boldsymbol{p}=(1-p_{\tau})\boldsymbol{p}+ p_{\tau}\boldsymbol{e}$, where $\boldsymbol{e}$ is the one-hot vector with 1 in the $\tau-$th position.
In order to obtain the Pareto stationary point,  
CU iterates the process of finding the toughest client several times to obtain the best combination $\boldsymbol{p}^*$ that alleviates the inconsistent deviations of clients.
Finally, we find the consistent optimization direction with the Pareto optimal solution $\boldsymbol{p}^*$ for aggregating in Eq.~\eqref{eq:agg}.
% 
% Hence, we obtain the approximately consistent global optimum and local optimum from both the optimization objective and optimization direction.
% 

Given three main modules, i.e., HPTI, HPL, and CA, we illustrate the overall algorithm of modeling \modelname~in Algo.~\ref{alg:HyperFed}.
Steps 1-10 are the server execution.
In step2, the server initializes model parameters and HPTI in it initializes the hyperbolic class prototypes.
Then for each communication round, all clients use HPL to train their local model with the shared Poincar\'e ball predictor in step 6.
After that, in step 8, CA in server receives the model parameters of all clients, and mitigates the inconsistency of client deviations in aggregation.
The details of client execution is listed in steps 11-21.

% \xenia{Theorem1 communication error bound}
% \xenia{Theorem2 Convergence rounds in total}

% \nosection{Privacy-preserving}
% % P4: for privacy concerns, 加了差分隐私
% % 1. 差分隐私在user-level 的情况下的privacy bound 
% \subsection{Global Aggregation}
% P1: 为什么在server上进行聚合，具体的聚合操作是什么,聚合对PFL这个问题的好处是什么
%P2: Algorithm

%
% \input{chapters/analysis}
% 
\section{Experiments and Discussion}

% Questions

\subsection{Experimental Setup}
\paragraph{Datasets.}
% ref Personalized fedrated learning (self-FL)
% ref sampling from MOON
We use four public datasets in torchvision\footnote{https://pytorch.org/vision/stable/index.html}, i.e., EMNIST by Letters~\cite{cohen2017emnist}, Fashion-MNIST (FMNIST)~\cite{xiao2017fashion}, Cifar10, and Cifar100~\cite{krizhevsky2009learning}, which are widely-used in the recent FL work~\cite{chen2021bridging,oh2021fedbabu,li2021model}.
There are two evaluation goals for FL with non-IID data, i.e., global performance (G-FL) and local personalized performance (P-FL).
G-FL tests the global model aggregated in the server by evaluating the \textit{test set} published in the torchvision.
In P-FL, we simulate local data distribution using the \textit{train set} published in torchvision to evaluate the  local models.
For all datasets, we simulate the non-IID data distributions following \cite{hsu2019measuring,li2021model}. 
Specifically, we sample a proportion of instances of class $j$ to client $k$ with Dirichlet distribution, i.e., $p_{j,k} \sim {Dir}_N (\alpha)$,  where $\alpha$ denotes the non-IID degree of every class among the clients.
The smaller $\alpha$ indicates the more heterogeneous data distribution.
We 
sample 75\% of local data as \textit{local training set} and the remaining as \textit{local test set}.

\paragraph{Comparison Methods.}
% ref Federated Learning with Label Distribution Skew via Logits Calibration
% 
We compare \modelname~with three categories of state-of-the-art approaches according to their optimization goals, i.e., (1) optimizing global model: \textbf{FedAvg}~\cite{mcmahan2017communication}, \textbf{FedProx}~\cite{li2020federated}, \textbf{SCAFFOLD}~\cite{karimireddy2019scaffold}, \textbf{FedDYN}~\cite{acar2020federated}, \textbf{MOON}~\cite{li2021model}, (2) optimizing local personalized models: \textbf{FedMTL}~\cite{smith2017federated}, \textbf{FedPer}~\cite{arivazhagan2019federated}, \textbf{pFedMe}~\cite{t2020personalized}, \textbf{Ditto}~\cite{li2021ditto},
\textbf{APPLE}~\cite{luo2021adapt}, and (3) optimizing both global and local models: \textbf{Fed-RoD}~\cite{chen2021bridging}, \textbf{FedBABU}~\cite{oh2021fedbabu}, and \textbf{SphereFed}~\cite{dong2022spherefed}. 
% 
% We provide the details of these methods in Appendix B.

\paragraph{Implementation Details.} 
We adopt ConvNet~\cite{lecun1998gradient} as a feature extractor for EMNIST and FMNIST, while ResNet~\cite{he2016deep} for Cifar10 and Cifar100.
We set all of the datasets with batch size as 128 and embedding dimension as 20.
 For \modelname, we choose RSGD~\cite{bonnabel2013stochastic} as the optimizer, set the learning rate $lr=0.3$, the margin $m=3$, and the slope degree $s=0.9$.
% 
% 
% In our basic setting,
We conduct training for all of the methods with 5 local epochs per round until converge.
We evaluate both G-FL and P-FL by top-1 accuracy.
We set the non-IID degree $\alpha=\{0.1, 0.5,5\}$, respectively, for evaluating the performance of different methods.
% 
% We depict the data distributions in Appendix A.
% % 
% % 
% We tune the hyper-parameters of all methods to their best values for a fair comparison, which is detailed in Appendix B.

\begin{figure}[t]
\centering
\includegraphics[width=0.85\linewidth]{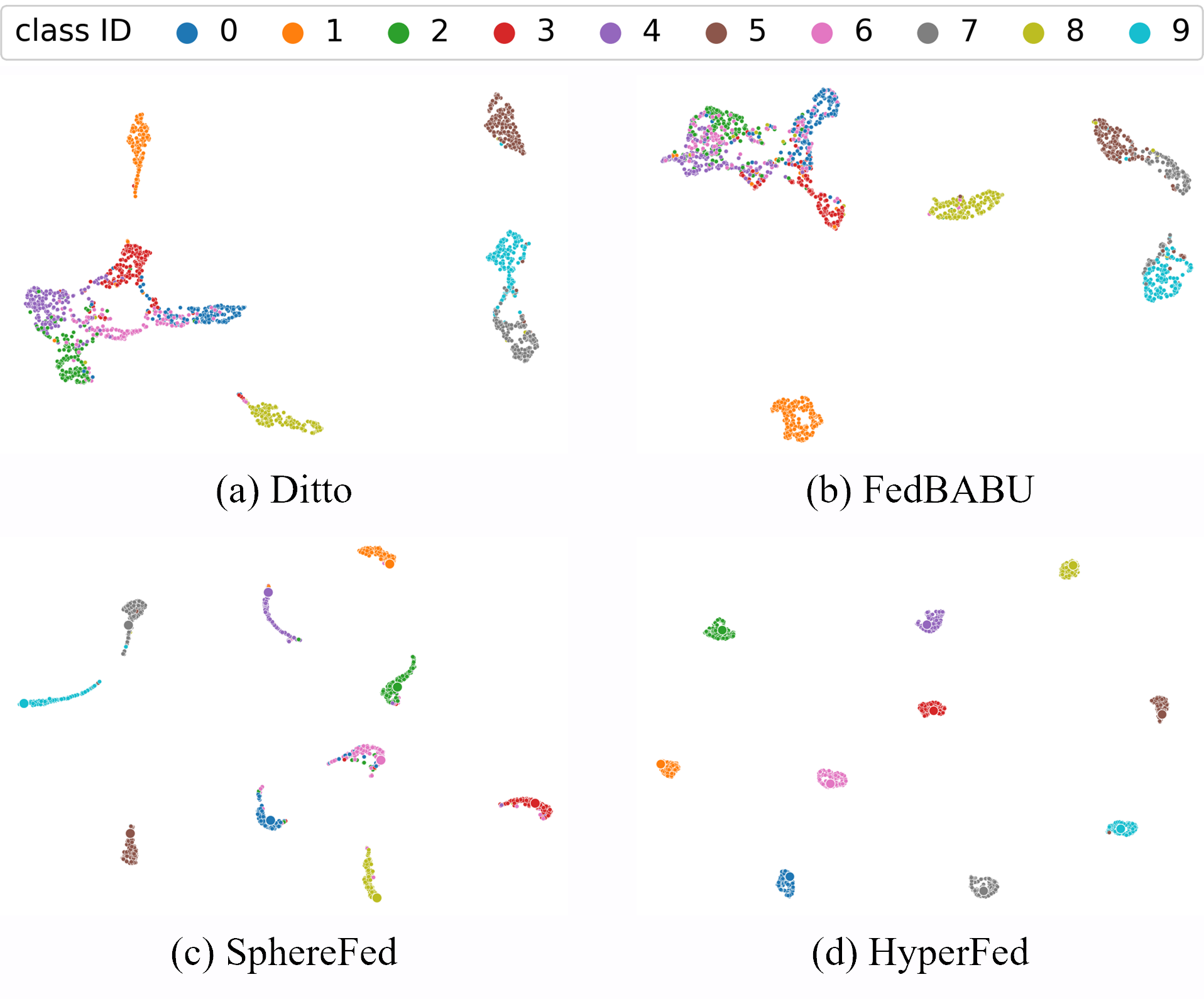}
\caption{UMAP visualizations on FMNIST ($\alpha=0.5$).}
\label{fig:visualization}
\end{figure}

\begin{table*}[]
\renewcommand\arraystretch{0.5}
\setlength\tabcolsep{6pt}
\centering
\resizebox{\textwidth}{!}{
\begin{tabular}{c|ccc|ccc|ccc|ccc}
\toprule
Dataset       & \multicolumn{3}{c}{EMNIST}   & \multicolumn{3}{c}{FMNIST}   & \multicolumn{3}{c}{Cifar10} & \multicolumn{3}{c}{Cifar100} \\
\midrule
Method \textbackslash NonIID        & Dir(0.1) & Dir(0.5) & Dir(5) & Dir(0.1) & Dir(0.5) & Dir(5) & Dir(0.1) & Dir(0.5) & Dir(5) & Dir(0.1)  & Dir(0.5) & Dir(5) \\
\midrule
FedAvg & 88.86  & 92.86  & 93.31  & 78.51  & 86.75  & 88.75  & 36.82  & 62.44  & 67.53  & 27.61  & 29.26  & 30.12  \\
FedProx                        & 90.75   & 93.27   & 93.65   & 78.49   & 86.57   & 88.68   & 37.95   & 63.48   & 67.23   & 26.91   & 29.77   & 29.98   \\
SCAFFOLD                       & 90.35   & 93.38   & 93.70   & 79.20   & 87.03   & 88.80   & 33.10   & 66.99   & 70.53   & 29.57   & 33.25   & 33.86    \\
FedDYN                         & 91.30   & 92.63   & 93.13   & 85.25   & 89.46   & 90.51   & 35.16   & 65.07   & 69.04   & 29.16   & 31.49   & 32.21   \\
MOON                           & 91.97   & 93.50   & 93.91   & 83.78   & 90.27   & 91.08   & 33.54   & 60.23   & 62.45   & 22.86   & 24.49   & 25.99   \\
\midrule
Fed-RoD                        & 89.42  & 92.80  & 93.52  & 77.26  & 89.17  & 90.78  & 37.79  & 66.90  & 70.81  & 17.06  & 24.32  & 31.99  \\
FedBABU & 86.34 & 91.95 & 92.74 & 74.31 & 82.44 & 85.21 & 37.90 & 60.11 & 65.16 & 22.98 & 24.56 & 23.89 
\\
SphereFed                      & 93.25   & 93.93   & 94.07   & 88.55   & 90.55   & 91.17   & 32.41   & 70.02   & 70.13   & 22.37   & 24.96   & 24.72   \\
\midrule
\modelname-\texttt{Geodesic}          & 93.35   & 94.13   & 94.43   & 79.00   & 90.76   & 91.48   & 34.26   & 65.53   & 68.65   & 26.05   & 29.07   & 26.62   \\
% \modelname-$\mathbb{E}$                     & 92.81   & 93.74   & 94.34   & 84.34   & 90.63   & 91.09   & 37.22   & 68.48   & 71.94   & 28.87   & 28.25   & 27.37   \\
\modelname-\texttt{Shared}                & 93.82   & 93.98   & 94.16   & 85.29   & 90.48   & 91.23   & 36.44   & 69.21   & 73.16   & 33.62   & 35.31   & 36.30   \\
\modelname-\texttt{Averaged}              & 93.65   & 94.29   & 94.45   & 87.91   & 90.98   & 91.69   & 36.69   & 70.12   & 70.81   & 33.39   & 36.33   & 36.93   \\
\modelname & \textbf{94.00}     &   \textbf{94.33}    &   \textbf{94.46} &   \textbf{89.38}   &   \textbf{91.16}   &   \textbf{91.83} & \textbf{38.03}   &   \textbf{71.25}   &   \textbf{75.22} &   \textbf{33.93}    &   \textbf{38.89}   &   \textbf{37.50}\\
 \bottomrule
\end{tabular}
}
\caption{G-FL accuracy ($\%$) of the global model. We bold  the best result.}
\label{tb:G-FL}
\end{table*}

\begin{table*}[]

\renewcommand\arraystretch{0.5}
\setlength\tabcolsep{6pt}
\centering
\resizebox{\textwidth}{!}{
\begin{tabular}{c|ccc|ccc|ccc|ccc}
\toprule
Dataset       & \multicolumn{3}{c}{EMNIST}    & \multicolumn{3}{c}{FMNIST}    & \multicolumn{3}{c}{Cifar10}  & \multicolumn{3}{c}{Cifar100} \\
\midrule
Method \textbackslash NonIID        & Dir(0.1) & Dir(0.5) & Dir(5)  & Dir(0.1) & Dir(0.5) & Dir(5)  & Dir(0.1) & Dir(0.5) & Dir(5)  & Dir(0.1) & Dir(0.5) & Dir(5)  \\
\midrule
FedMTL                         & 96.28  & 91.08  & 88.21  & 97.39  & 90.89  & 85.71  & 90.57  & 65.68  & 48.31  & 46.28  & 24.72  & 11.64  \\
FedPer                         & 97.23  & 94.19  & 92.97  & 96.87  & 90.54  & 87.21  & 91.93  & 75.63  & 67.20  & 49.69  & 32.75  & 23.89  \\
pFedMe                         & 97.23  & 94.03  & 92.62  & 96.03  & 88.57  & 84.95  & 92.45  & 77.64  & 66.09  & 55.20  & 36.34  & 28.19  \\
Ditto                          & 97.72  & 95.32  & 94.30  & 97.79  & 93.75  & 92.05  & 91.34  & 74.35  & 69.94  & 47.08  & 31.89  & 27.04  \\
APPLE    &
97.19  & 94.08  & 92.63  & 96.79  & 90.49  & 86.59  & 89.85  & 67.96  & 56.13  & 42.94  & 24.61  & 20.39  \\
\midrule
Fed-RoD                        & 97.76  & 95.31  & 93.99  & 97.21  & 93.48  & 91.62  & 91.24  & 72.11  & 71.33  & 35.03  & 27.70  & 32.22  \\
FedBABU                        & 97.36  & 94.17  & 92.81  & 96.64  & 89.59  & 85.57  & 92.04  & 64.05  & 63.46  & 29.30  & 25.05  & 24.01  \\
SphereFed     & 93.61  & 94.11  & 94.34  & 88.91  & 90.87  & 91.88  & 91.68  & 78.39  & 72.86  & 40.68 & 32.63  & 34.90    \\
\midrule
\modelname-\texttt{Geodesic}          & 97.97   & 95.60   & 94.48   & 97.00  & 93.98   & 91.89   & 91.68   & 72.20   & 69.02   & 29.70   & 27.11   & 27.03   \\
% \modelname-$\mathbb{E}$                     & 92.80   & 93.82   & 93.99   & 84.88   & 90.72   & 91.80   & 92.66   & 76.55   & 69.67   & 27.73   & 25.76   & 26.18   \\
\modelname-\texttt{Fixed}                 & 97.96   & 95.65   & 94.45   & 96.53   & 92.66   & 90.85   & 90.57   & 79.66   & 59.38   & 49.08   & 20.20   & 11.31   \\
\modelname-\texttt{Shared}                & 97.50   & 95.66   & 94.39   & 97.50   & 93.00   & 91.52   & 92.69   & 77.43   & 71.62   & 47.97   & 33.16   & 35.19   \\
\modelname-\texttt{Averaged}              & 97.71   & 95.13   & 94.46   & 97.83   & 93.73   & 91.96  & 92.63   & 70.53    & 69.02    & 53.50   & 36.49    & 35.79    \\
\modelname & \textbf{98.76}  & \textbf{96.11} & \textbf{94.53}  & \textbf{98.13}  & \textbf{94.19}  & \textbf{92.08}  & \textbf{93.49} & \textbf{83.17}  & \textbf{74.94}  & \textbf{59.85} & \textbf{40.94}  & \textbf{36.80}  \\
\bottomrule
\end{tabular}
}
\caption{P-FL accuracy ($\%$) of the local model. We bold  the best result.}
\label{tb:P-FL}
\end{table*}

\begin{figure}[t]
\centering
\subfigure[G-FL]{
\begin{minipage}{0.47\columnwidth}{
\centering
\label{fig:E-pFL-cifar10-0.5}
\includegraphics[scale=0.5]{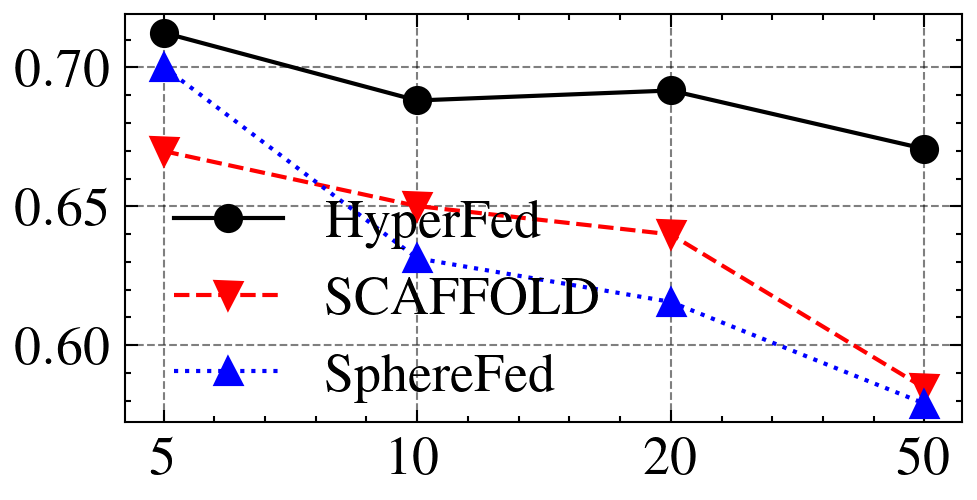} %
}
\end{minipage}
}
\subfigure[P-FL]{
\begin{minipage}{0.47\columnwidth}{
\centering
\label{fig:E-PFL-cifar10-0.5}
\includegraphics[scale=0.5]{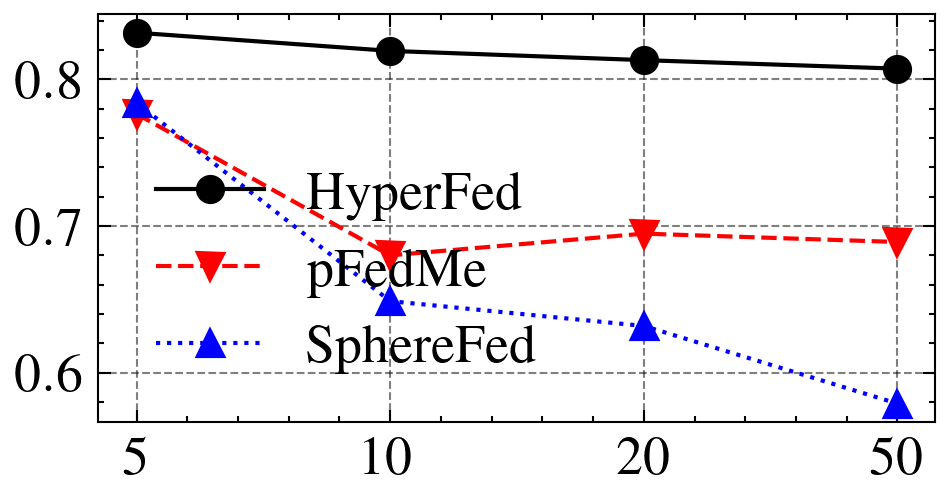} % 
}
\end{minipage}
}
\caption{Effect of local epochs $E$ on Cifar10 ($\alpha=0.5$).}
\label{fig:hyperparameter_E}
\end{figure}

\begin{figure}[t]
\centering
\subfigure[G-FL]{
\begin{minipage}{0.47\columnwidth}{
\centering
\label{fig:cifar10_s_GFL}
\includegraphics[scale=0.5]{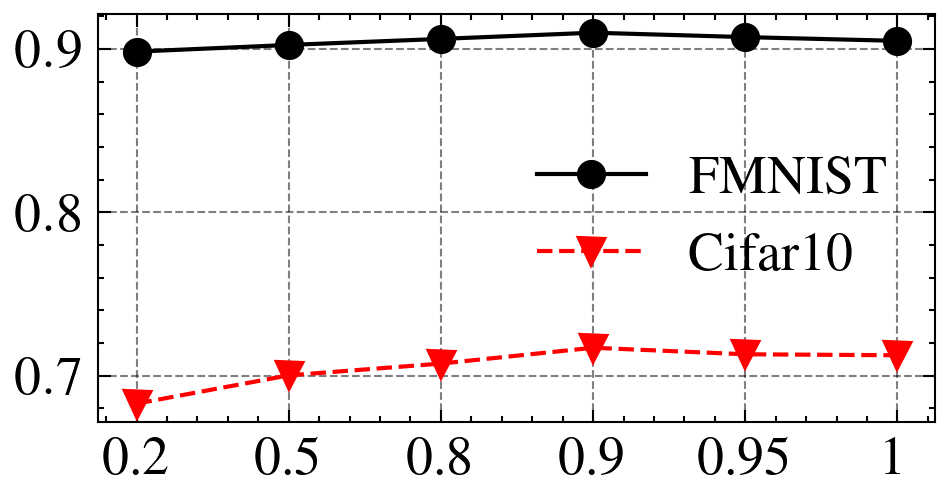} %
}
\end{minipage}
}
\subfigure[P-FL]{
\begin{minipage}{0.47\columnwidth}{
\centering
\label{fig:cifar10_s_PFL}
\includegraphics[scale=0.5]{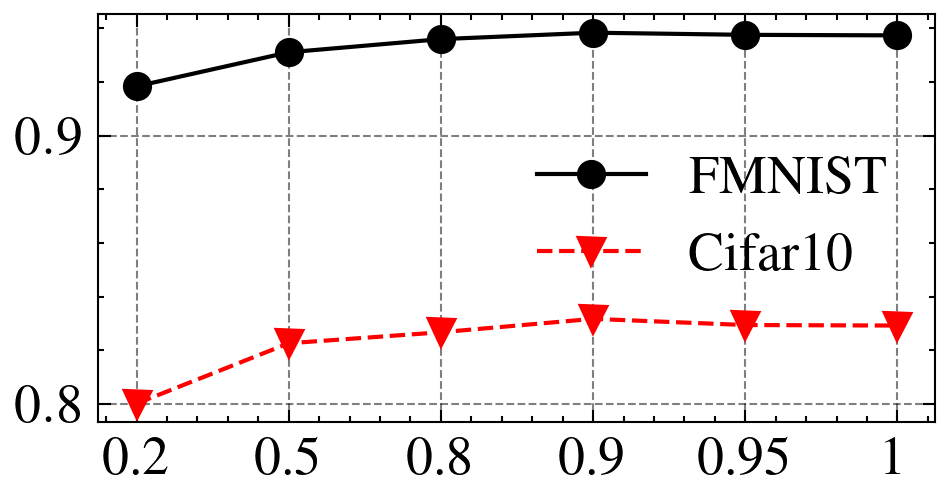} % 
}
\end{minipage}
}
\caption{Effect of slope degree $s$ on FMNIST $\&$ Cifar10 ($\alpha=0.5$)}
\label{fig:slope-degree}
\end{figure}

\begin{figure}[t]
\centering
\subfigure[G-FL]{
\begin{minipage}{0.47\columnwidth}{
\centering
\label{fig:cifar10_m_GFL}
\includegraphics[scale=0.5]{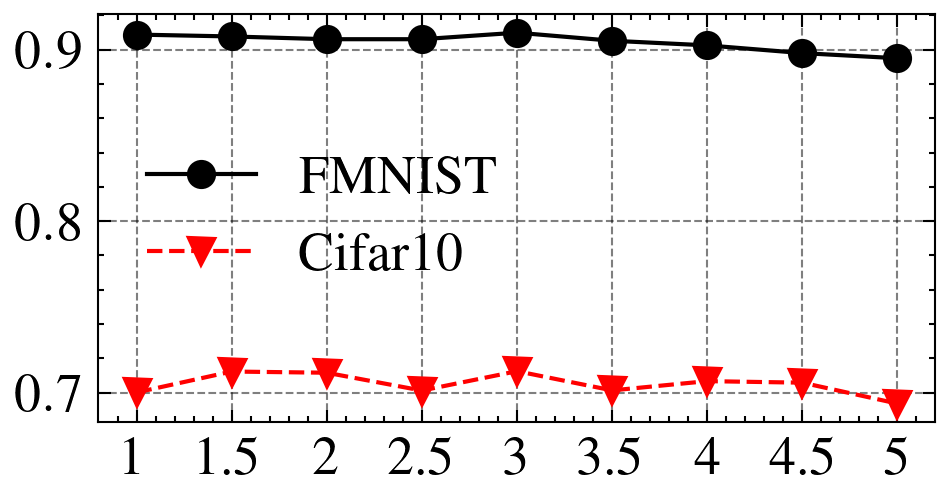} %
}
\end{minipage}
}
\subfigure[P-FL]{
\begin{minipage}{0.47\columnwidth}{
\centering
\label{fig:cifar10_m_PFL}
\includegraphics[scale=0.5]{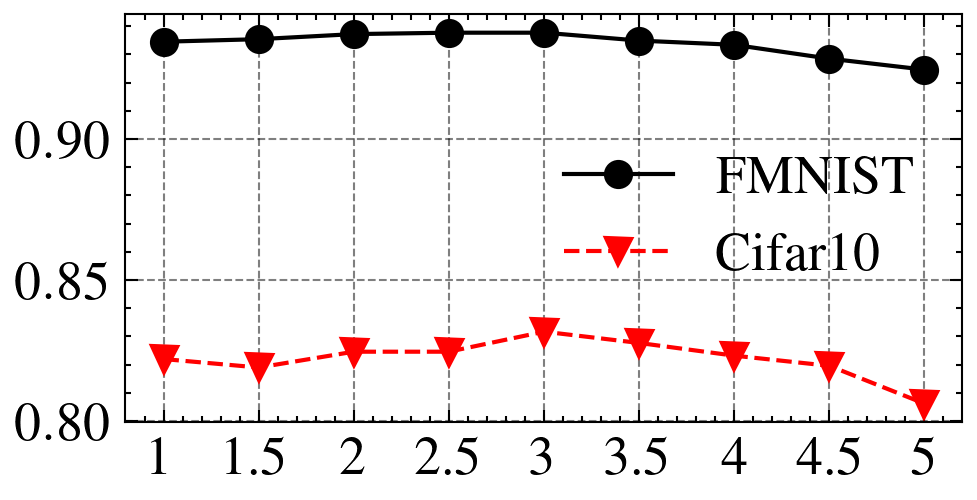} % 
}
\end{minipage}
}
\caption{Effect of margin $m$ on FMNIST $\&$ Cifar10 ($\alpha=0.5$).}
\label{fig:margin_hyper}
\end{figure}

\begin{figure}[t]
\centering
\subfigure[G-FL]{
\begin{minipage}{0.47\columnwidth}{
\centering
\label{fig:fmnist_n_GFL}
\includegraphics[scale=0.5]{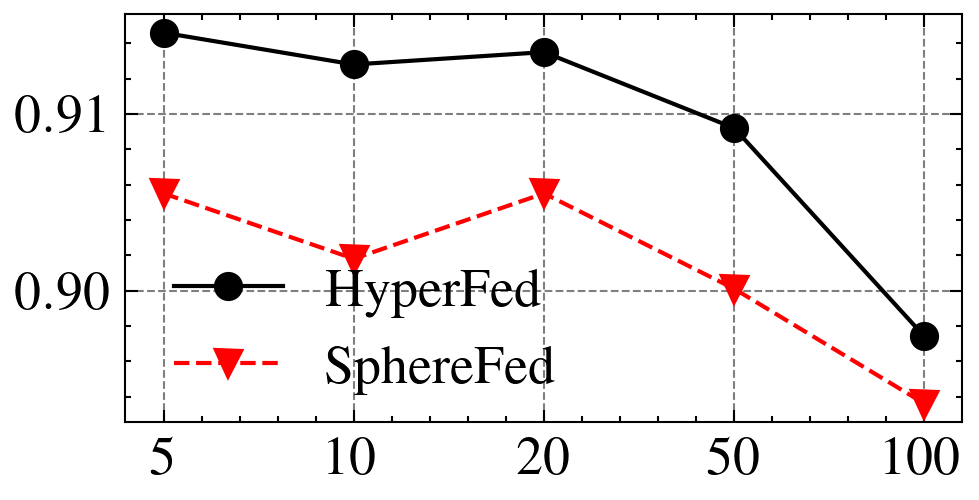} %
}
\end{minipage}
}
\subfigure[P-FL]{
\begin{minipage}{0.47\columnwidth}{
\centering
\label{fig:fmnist_n_PFL}
\includegraphics[scale=0.5]{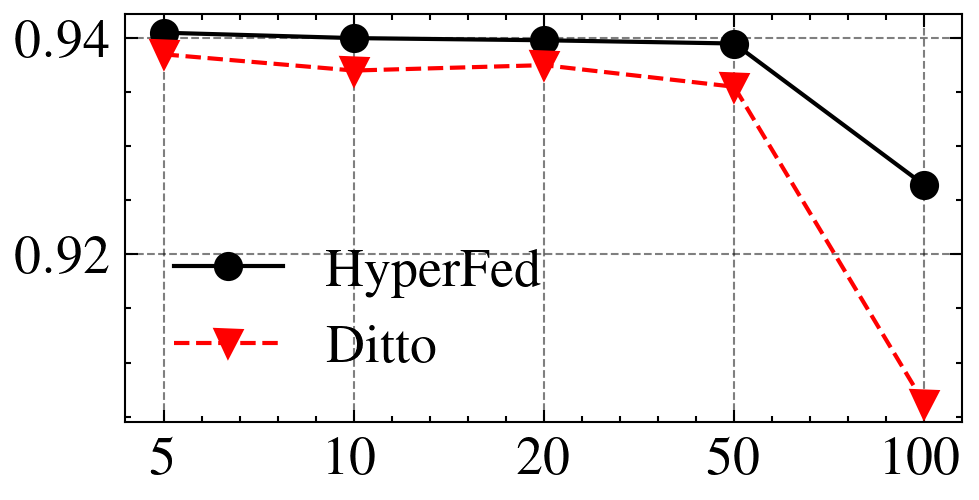} % 
}
\end{minipage}
}
\caption{Effect of clients number $K$ on FMNIST ($\alpha=0.5$). }
\label{fig:n_clients}
\end{figure}

% Please add the following required packages to your document preamble:
\begin{table}[]
\renewcommand\arraystretch{0.5}
\setlength\tabcolsep{6pt}
\centering
\resizebox{\columnwidth}{!}{
\begin{tabular}{cccccccccc}
\toprule
Dataset  & Goal & 2       & 5       & 10      & 20      & 25      & 50      & 100     \\
\midrule
FMNIST
% \multirow{2}{*}{ } 
& P-FL & 88.91  & 93.20  & 93.84  & 93.98  & 93.70  & 93.73  & 93.66  \\
 ($\alpha$=0.5)       & G-FL    & 88.59  & 91.07  & 90.91  & 91.16 & 90.74  & 90.74  & 90.74  \\
\midrule
Cifar10
% \multirow{2}{*}{} 
& P-FL & 72.82  & 80.55  & 82.83  & 83.17  & 82.50  & 83.15  & 82.76  \\
  ($\alpha$=0.5)        & G-FL    & 60.87  & 68.82  & 69.84  & 71.25  & 70.67  & 71.51  & 71.69 
        \\
\bottomrule
\end{tabular}
}
\caption{Effect of the representation dimension in \modelname.}
\label{tb:dimension}
\end{table}

\subsection{Empirical Results}

\paragraph{Performance Comparison.}
We run each method 5 times and report the performance of G-FL and P-FL in Tab.~\ref{tb:G-FL}-\ref{tb:P-FL} with the average value, respectively.
There are mainly three observations.
(1) \textbf{In terms of G-FL} evaluated in Tab.~\ref{tb:G-FL}, generally speaking, the performance of G-FL is increasing along with the increase of non-IID degree $\alpha$.
It states that the heterogeneity of data distribution deteriorates the performance of FL methods.
FedAvg takes no measure to handle non-IID, which is worse than most methods.
Similarly, FedBABU is a variant of FedAvg that fixes the randomly initialized classifier, perform inferior as well.
It means that simply fixing the classifier initialized by random is not enough for FL with non-IID data.
% 
% MOON is effective on small EMNIST and FMNIST, but they have much worse accuracy on Cifar10 and Cifar100.
% % 
% FedDYN and Fed-ROD maintain comparable results on four datasets.
% % 
% FedBABU is less effective, which means that simply fixing the classifier initialized by random is not enough for FL with non-IID data.
%
% SphereFed is the runner-up on EMNIST and FMNIST, but it fails to be effective in complicated datasets, i.e., Cifar10 and Cifar100.
% % 
% It states simply fixing the classifier head but not utilizing the hierarchical information, will hinder the scalability of FL methods for complicated datasets.
% % 
% SCAFFOLD is the runner-up on Cifar10 ($\alpha=0.5$ and $\alpha=5$) and Cifar100 ($\alpha=0.5$ and $\alpha=5$), which means it can handle complicated datasets with less heterogeneous data distribution.
% % 
(2) \textbf{In terms of P-FL} evaluated in Tab.~\ref{tb:P-FL},
the methods of the third category can achieve accuracy similar as the second category on FMNIST and EMNIST, but cannot maintain the personalization on Cifar100.
This indicates that simply fixing class statistics in FL is  less promising than personalization strategies on complex dataset. 
SphereFed is the runner-up of Cifar100 ($\alpha=5$), but it fails to perform well on more heterogeous setting, i.e., Cifar100 ($\alpha=\{0.1, 0.5\}$).
This phenomenon shows that \modelname~performs better for FL with non-IID data, as a FL method of the third category.
% 
% 
% (3) \textbf{In terms of the performance of methods simultaneously optimizing the local and global model}, FedBABU and SphereFed achieve inferior performance on Cifar10 and Cifar100 than global model personalization methods in G-FL and learning personalized model in P-FL.
% 
% This indicates that naively fixing the classifier head will hinder the scalability of FL methods for complicated datasets.
(3) \textbf{In terms of the performance of \modelname}, \modelname~achieves the best results in all kinds of non-IID degrees and different datasets, which verifies the efficacy of \modelname~for non-IID problems. 
% 
% Compared with EMNIST and FMNIST, 
% \modelname~achieves fairly more performance enhancement on Cifar10 and Cifar100.
% % % 
% It states the necessity of improving the sufficient representation for \xenia{complex datasets}.
% better PFL performance.
% % 
 In the results of Cifar100,
\modelname~significantly outperforms all of the comparison algorithms listed both in G-FL and P-FL, in terms of at least 10.75\%  and 5.44\%, respectively.
This shows that \modelname~can capture more fine-grained representation to improve performance, especially on large and complicated datasets with low dimensional representation.

\noindent\textbf{Visualization. }
To verify the benefits of fixing class statistics in hyperbolic space, we utilize UMAP~\cite{mcinnes2018umap} to visualize the G-FL hidden representations of the global model in \modelname, the runner-up of G-FL,i.e., SphereFed, and P-FL, i.e., Ditto, and FedBABU on FMNIST ($\alpha=0.5$) in Fig.~\ref{fig:visualization}. 
we can find that: (1) Compared with Ditto, methods fixing the class statistics attain more gathered representations.
(2) Though FedBABU fixes the classifier, the randomly initialized classifier is limited in resolving class overlapping.
(3) Compared with SphereFed whose classifier is initialized with a set of orthogonal basis in hypersphere,  \modelname~concentrates the hidden representation tighter due to sufficiently utilizing the hierarchical information.

\paragraph{Ablation Studies.}
% 
% hyperbolic vs Euclidean 
% triplet -> representation no hyperbolic
% triplet for euclidean
% 
We consider four variants of \modelname: 
% (1). \modelname~using fixed and shared prototypes with contrastive learning in Euclidean space (Cosine Distance), i.e., \modelname-$\mathbb{E}$,
(1) \modelname~uses geodesic
% (Eq.~\eqref{eq:geo_dis}) 
as metric, i.e., \modelname-\texttt{Geodesic},
(2) \modelname~uses fixed class prototypes only, i.e., \modelname-\texttt{fixed}, (3) \modelname~uses shared class prototypes only, i.e., \modelname-\texttt{shared}, and (4) \modelname~uses weighted average aggregation by data amounts,
% using the amounts of data as weights to average aggregation,
i.e., \modelname-\texttt{Averaged}.
% 
% The former is built by substituting the triplet loss in Eq.~\eqref{eq:loss} with the original geodesic distance in Eq.~\eqref{eq:geo_dis}.
% % 
% The latter is a Euclidean version of \modelname, i.e., a federated learning framework optimizes cosine distance with fixed Euclidean classifier. 
% 
% 
From Tab.~\ref{tb:G-FL}-\ref{tb:P-FL}, we can discover that:
All of the variants decrease their performance compared with \modelname.
These results identify that all of the components of \modelname, 
% i.e., hyperbolic triplet loss, fixed and shared class prototypes, and consistent updating, 
contribute to performance enhancement.

\noindent\textbf{Hyper-parameters Sensitivity.}
To study the sensitivity of hyper-parameters, we compare the performance of FMNIST and Cifar10 by varying the local epochs $E= \{5, 10,20, 50\}$ in Fig.~\ref{fig:hyperparameter_E}, slope degree $s=\{0.2, 0.5, 0.8, 0.9, 0.95, 1\}$ in Fig.~\ref{fig:slope-degree}, margin $m=\{1, 1.5, 2,2.5, 3, 3.5, 4, 4.5, 5\}$ in Fig.~\ref{fig:margin_hyper}, number of clients $K= \{5, 10, 20,50,100\}$ in Fig.~\ref{fig:n_clients}, and the dimension of features representation $d=\{2,5,10,20,25,50,100\}$ in Tab.~\ref{tb:dimension},  respectively.
We can conclude that:
(1) The increase of local epochs per communication round decreases the performance of FL methods. 
(2) \modelname~nearly converges when $s=0.9$, which indicates that positioning the class prototypes in 90\% of radius achieves a balance between the best model performance and the most general class information. 
% 
% We evaluate the behavior of impacts of local epochs and find the performance is shrinking with additionally more epochs per round.
% 
(3) The performance of \modelname~varying by $m$ forms a slight bell curve.
% , which means $m=3$ contributes to best performance.
% 
(4) \modelname~outperforms the runner-up methods in all cases of different number of clients.
(5) The dimension changes slightly affect the performance of \modelname, proving that \modelname~is proficient in low-dimensional representation. 
% 

% local epochs 
% dim of model
% reg constant (e.g., margin)
% clients number
% model architecture
%

\section{Conclusion}
    To enhance federated learning (FL) with non-IID data, we propose \modelname~which contains hyperbolic prototype Tammes initialization (HPTI) module, hyperbolic prototype learning (HPL) module, and consistent aggregation (CA) module.
    Firstly, HPTI constructs uniformly distributed and fixed class prototypes on server, and
    shares them with clients to guide consistent feature representation for local clients.
    secondly, HPL models client data in the hyperbolic model space with the supervision of shared class prototypes.
    Additionally, CA mitigates the impact of inconsistent deviations from clients to server. 
    Extensive studies on four datasets prove the effectiveness of \modelname.
    % that \modelname~is effective in enhancing the performance of FL with non-IID data.
    % under the non-IID setting.
    % 

\section*{Acknowledgements}
This work was supported in part by the ``Pioneer” and ``Leading Goose” R\&D Program of Zhejiang (No.2022C01126), and Leading Expert of “Ten Thousands Talent Program” of Zhejiang Province (No.2021R52001).
% This work was supported in part by the ``Pioneer" and ``Leading Goose" R\&D Program of Zhejiang (No. 2022C01126), the National Natural Science Foundation of China (No. 62172362).
% \xenia{ant}

\bibliographystyle{named}
\bibliography{ijcai23}

\begin{thebibliography}{}

\bibitem[\protect\citeauthoryear{Acar \bgroup \em et al.\egroup
  }{2020}]{acar2020federated}
Durmus Alp~Emre Acar, Yue Zhao, Ramon Matas, Matthew Mattina, Paul Whatmough,
  and Venkatesh Saligrama.
\newblock Federated learning based on dynamic regularization.
\newblock In {\em ICLR}, 2020.

\bibitem[\protect\citeauthoryear{Arivazhagan \bgroup \em et al.\egroup
  }{2019}]{arivazhagan2019federated}
Manoj~Ghuhan Arivazhagan, Vinay Aggarwal, Aaditya~Kumar Singh, and Sunav
  Choudhary.
\newblock Federated learning with personalization layers.
\newblock {\em arXiv preprint arXiv:1912.00818}, 2019.

\bibitem[\protect\citeauthoryear{Bonnabel}{2013}]{bonnabel2013stochastic}
Silvere Bonnabel.
\newblock Stochastic gradient descent on riemannian manifolds.
\newblock {\em IEEE Transactions on Automatic Control}, 58(9):2217--2229, 2013.

\bibitem[\protect\citeauthoryear{Chen and Chao}{2021}]{chen2021bridging}
Hong-You Chen and Wei-Lun Chao.
\newblock On bridging generic and personalized federated learning for image
  classification.
\newblock In {\em ICLR}, 2021.

\bibitem[\protect\citeauthoryear{Chen \bgroup \em et al.\egroup
  }{2022a}]{CCIjcai22}
Chaochao Chen, Jun Zhou, Longfei Zheng, Huiwen Wu, Lingjuan Lyu, Jia Wu,
  Bingzhe Wu, Ziqi Liu, Li~Wang, and Xiaolin Zheng.
\newblock Vertically federated graph neural network for privacy-preserving node
  classification.
\newblock In {\em IJCAI}, pages 1959--1965. ijcai.org, 2022.

\bibitem[\protect\citeauthoryear{Chen \bgroup \em et al.\egroup
  }{2022b}]{chen2022fully}
Weize Chen, Xu~Han, Yankai Lin, Hexu Zhao, Zhiyuan Liu, Peng Li, Maosong Sun,
  and Jie Zhou.
\newblock Fully hyperbolic neural networks.
\newblock In {\em Proceedings of the 60th Annual Meeting of the Association for
  Computational Linguistics}, pages 5672--5686, 2022.

\bibitem[\protect\citeauthoryear{Cho \bgroup \em et al.\egroup
  }{2019}]{cho2019large}
Hyunghoon Cho, Benjamin DeMeo, Jian Peng, and Bonnie Berger.
\newblock Large-margin classification in hyperbolic space.
\newblock In {\em PMLR}, pages 1832--1840. PMLR, 2019.

\bibitem[\protect\citeauthoryear{Cohen \bgroup \em et al.\egroup
  }{2017}]{cohen2017emnist}
Gregory Cohen, Saeed Afshar, Jonathan Tapson, and Andre Van~Schaik.
\newblock Emnist: Extending mnist to handwritten letters.
\newblock In {\em IJCNN}, pages 2921--2926. IEEE, 2017.

\bibitem[\protect\citeauthoryear{D{\'e}sid{\'e}ri}{2012}]{desideri2012multiple}
Jean-Antoine D{\'e}sid{\'e}ri.
\newblock Multiple-gradient descent algorithm (mgda) for multiobjective
  optimization.
\newblock {\em Comptes Rendus Mathematique}, 350(5-6):313--318, 2012.

\bibitem[\protect\citeauthoryear{Dong \bgroup \em et al.\egroup
  }{2022}]{dong2022spherefed}
Xin Dong, Sai~Qian Zhang, Ang Li, and HT~Kung.
\newblock Spherefed: Hyperspherical federated learning.
\newblock In {\em ECCV}, pages 165--184. Springer, 2022.

\bibitem[\protect\citeauthoryear{Ermolov \bgroup \em et al.\egroup
  }{2022}]{ermolov2022hyperbolic}
Aleksandr Ermolov, Leyla Mirvakhabova, Valentin Khrulkov, Nicu Sebe, and Ivan
  Oseledets.
\newblock Hyperbolic vision transformers: Combining improvements in metric
  learning.
\newblock In {\em CVPR}, pages 7409--7419, 2022.

\bibitem[\protect\citeauthoryear{Fallah \bgroup \em et al.\egroup
  }{2020}]{fallah2020personalized}
Alireza Fallah, Aryan Mokhtari, and Asuman Ozdaglar.
\newblock Personalized federated learning with theoretical guarantees: A
  model-agnostic meta-learning approach.
\newblock {\em NeurIPS}, 33:3557--3568, 2020.

\bibitem[\protect\citeauthoryear{Ganea \bgroup \em et al.\egroup
  }{2018}]{ganea2018hyperbolic}
Octavian Ganea, Gary B{\'e}cigneul, and Thomas Hofmann.
\newblock Hyperbolic entailment cones for learning hierarchical embeddings.
\newblock In {\em ICML}, pages 1646--1655. PMLR, 2018.

\bibitem[\protect\citeauthoryear{Ghadimi~Atigh \bgroup \em et al.\egroup
  }{2021}]{ghadimi2021hyperbolic}
Mina Ghadimi~Atigh, Martin Keller-Ressel, and Pascal Mettes.
\newblock Hyperbolic busemann learning with ideal prototypes.
\newblock {\em NeurIPS}, 34:103--115, 2021.

\bibitem[\protect\citeauthoryear{He \bgroup \em et al.\egroup
  }{2016}]{he2016deep}
Kaiming He, Xiangyu Zhang, Shaoqing Ren, and Jian Sun.
\newblock Deep residual learning for image recognition.
\newblock In {\em CVPR}, pages 770--778, 2016.

\bibitem[\protect\citeauthoryear{Hsu \bgroup \em et al.\egroup
  }{2019}]{hsu2019measuring}
Tzu-Ming~Harry Hsu, Hang Qi, and Matthew Brown.
\newblock Measuring the effects of non-identical data distribution for
  federated visual classification.
\newblock {\em arXiv preprint arXiv:1909.06335}, 2019.

\bibitem[\protect\citeauthoryear{Jaggi}{2013}]{jaggi2013revisiting}
Martin Jaggi.
\newblock Revisiting frank-wolfe: Projection-free sparse convex optimization.
\newblock In {\em ICML}, pages 427--435. PMLR, 2013.

\bibitem[\protect\citeauthoryear{Karimireddy \bgroup \em et al.\egroup
  }{2019}]{karimireddy2019scaffold}
Sai~Praneeth Karimireddy, Satyen Kale, Mehryar Mohri, Sashank~J Reddi,
  Sebastian~U Stich, and Ananda~Theertha Suresh.
\newblock Scaffold: Stochastic controlled averaging for on-device federated
  learning.
\newblock In {\em ICML}. PMLR, 2019.

\bibitem[\protect\citeauthoryear{Khrulkov \bgroup \em et al.\egroup
  }{2020}]{khrulkov2020hyperbolic}
Valentin Khrulkov, Leyla Mirvakhabova, Evgeniya Ustinova, Ivan Oseledets, and
  Victor Lempitsky.
\newblock Hyperbolic image embeddings.
\newblock In {\em CVPR}, pages 6418--6428, 2020.

\bibitem[\protect\citeauthoryear{Krizhevsky \bgroup \em et al.\egroup
  }{2009}]{krizhevsky2009learning}
Alex Krizhevsky, Geoffrey Hinton, et~al.
\newblock Learning multiple layers of features from tiny images.
\newblock {\em Technical Report, University of Toronto}, 2009.

\bibitem[\protect\citeauthoryear{LeCun \bgroup \em et al.\egroup
  }{1998}]{lecun1998gradient}
Yann LeCun, L{\'e}on Bottou, Yoshua Bengio, and Patrick Haffner.
\newblock Gradient-based learning applied to document recognition.
\newblock {\em Proceedings of the IEEE}, 86(11):2278--2324, 1998.

\bibitem[\protect\citeauthoryear{Li \bgroup \em et al.\egroup
  }{2019}]{li2019convergence}
Xiang Li, Kaixuan Huang, Wenhao Yang, Shusen Wang, and Zhihua Zhang.
\newblock On the convergence of fedavg on non-iid data.
\newblock {\em arXiv preprint arXiv:1907.02189}, 2019.

\bibitem[\protect\citeauthoryear{Li \bgroup \em et al.\egroup
  }{2020}]{li2020federated}
Tian Li, Anit~Kumar Sahu, Manzil Zaheer, Maziar Sanjabi, Ameet Talwalkar, and
  Virginia Smith.
\newblock Federated optimization in heterogeneous networks.
\newblock {\em Proceedings of Machine Learning and Systems}, 2:429--450, 2020.

\bibitem[\protect\citeauthoryear{Li \bgroup \em et al.\egroup
  }{2021a}]{li2021model}
Qinbin Li, Bingsheng He, and Dawn Song.
\newblock Model-contrastive federated learning.
\newblock In {\em CVPR}, pages 10713--10722, 2021.

\bibitem[\protect\citeauthoryear{Li \bgroup \em et al.\egroup
  }{2021b}]{li2021ditto}
Tian Li, Shengyuan Hu, Ahmad Beirami, and Virginia Smith.
\newblock Ditto: Fair and robust federated learning through personalization.
\newblock In {\em ICML}, pages 6357--6368. PMLR, 2021.

\bibitem[\protect\citeauthoryear{Linial \bgroup \em et al.\egroup
  }{1995}]{linial1995geometry}
Nathan Linial, Eran London, and Yuri Rabinovich.
\newblock The geometry of graphs and some of its algorithmic applications.
\newblock {\em Combinatorica}, 15(2):215--245, 1995.

\bibitem[\protect\citeauthoryear{Liu \bgroup \em et al.\egroup
  }{2019}]{liu2019hyperbolic}
Qi~Liu, Maximilian Nickel, and Douwe Kiela.
\newblock Hyperbolic graph neural networks.
\newblock {\em NeurIPS}, 32, 2019.

\bibitem[\protect\citeauthoryear{Liu \bgroup \em et al.\egroup
  }{2020}]{liu2020hyperbolic}
Shaoteng Liu, Jingjing Chen, Liangming Pan, Chong-Wah Ngo, Tat-Seng Chua, and
  Yu-Gang Jiang.
\newblock Hyperbolic visual embedding learning for zero-shot recognition.
\newblock In {\em CVPR}, pages 9273--9281, 2020.

\bibitem[\protect\citeauthoryear{Liu \bgroup \em et al.\egroup
  }{2021}]{weimingNeurips21}
Weiming Liu, Jiajie Su, Chaochao Chen, and Xiaolin Zheng.
\newblock Leveraging distribution alignment via stein path for cross-domain
  cold-start recommendation.
\newblock In {\em NeurIPS}, pages 19223--19234, 2021.

\bibitem[\protect\citeauthoryear{Liu \bgroup \em et al.\egroup
  }{2023}]{weimingWWW23}
Weiming Liu, Xiaolin Zheng, Chaochao Chen, Jiajie Su, Xinting Liao, Mengling
  Hu, and Yanchao Tan.
\newblock Joint internal multi-interest exploration and external domain
  alignment for cross domain sequential recommendation.
\newblock In {\em WWW}, page 383–394, New York, NY, USA, 2023. Association
  for Computing Machinery.

\bibitem[\protect\citeauthoryear{Luo and Wu}{2021}]{luo2021adapt}
Jun Luo and Shandong Wu.
\newblock Adapt to adaptation: Learning personalization for cross-silo
  federated learning.
\newblock {\em arXiv preprint arXiv:2110.08394}, 2021.

\bibitem[\protect\citeauthoryear{Luo \bgroup \em et al.\egroup
  }{2022}]{luo2022disentangled}
Zhengquan Luo, Yunlong Wang, Zilei Wang, Zhenan Sun, and Tieniu Tan.
\newblock Disentangled federated learning for tackling attributes skew via
  invariant aggregation and diversity transferring.
\newblock In {\em ICML}, volume 162 of {\em Proceedings of Machine Learning
  Research}, pages 14527--14541. PMLR, 2022.

\bibitem[\protect\citeauthoryear{McInnes \bgroup \em et al.\egroup
  }{2018}]{mcinnes2018umap}
Leland McInnes, John Healy, and James Melville.
\newblock Umap: Uniform manifold approximation and projection for dimension
  reduction.
\newblock {\em arXiv preprint arXiv:1802.03426}, 2018.

\bibitem[\protect\citeauthoryear{McMahan \bgroup \em et al.\egroup
  }{2017}]{mcmahan2017communication}
Brendan McMahan, Eider Moore, Daniel Ramage, Seth Hampson, and Blaise~Aguera
  y~Arcas.
\newblock Communication-efficient learning of deep networks from decentralized
  data.
\newblock In {\em Artificial intelligence and statistics}, pages 1273--1282.
  PMLR, 2017.

\bibitem[\protect\citeauthoryear{Mettes \bgroup \em et al.\egroup
  }{2019}]{mettes2019hyperspherical}
Pascal Mettes, Elise van~der Pol, and Cees Snoek.
\newblock Hyperspherical prototype networks.
\newblock {\em NeurIPS}, 32, 2019.

\bibitem[\protect\citeauthoryear{Movshovitz-Attias \bgroup \em et al.\egroup
  }{2017}]{movshovitz2017no}
Yair Movshovitz-Attias, Alexander Toshev, Thomas~K Leung, Sergey Ioffe, and
  Saurabh Singh.
\newblock No fuss distance metric learning using proxies.
\newblock In {\em CVPR}, pages 360--368, 2017.

\bibitem[\protect\citeauthoryear{Nickel and Kiela}{2017}]{nickel2017poincare}
Maximillian Nickel and Douwe Kiela.
\newblock Poincar{\'e} embeddings for learning hierarchical representations.
\newblock {\em NeurIPS}, 30, 2017.

\bibitem[\protect\citeauthoryear{Oh \bgroup \em et al.\egroup
  }{}]{oh2021fedbabu}
Jaehoon Oh, SangMook Kim, and Se-Young Yun.
\newblock Fedbabu: Toward enhanced representation for federated image
  classification.
\newblock In {\em ICLR, 2021}.

\bibitem[\protect\citeauthoryear{Sekitani and
  Yamamoto}{1993}]{sekitani1993recursive}
Kazuyuki Sekitani and Yoshitsugu Yamamoto.
\newblock A recursive algorithm for finding the minimum norm point in a
  polytope and a pair of closest points in two polytopes.
\newblock {\em Mathematical Programming}, 61(1):233--249, 1993.

\bibitem[\protect\citeauthoryear{Sener and Koltun}{2018}]{sener2018multi}
Ozan Sener and Vladlen Koltun.
\newblock Multi-task learning as multi-objective optimization.
\newblock {\em Advances in neural information processing systems}, 31, 2018.

\bibitem[\protect\citeauthoryear{Shen \bgroup \em et al.\egroup
  }{2021}]{shen2021spherical}
Jiayi Shen, Zehao Xiao, Xiantong Zhen, and Lei Zhang.
\newblock Spherical zero-shot learning.
\newblock {\em IEEE Transactions on Circuits and Systems for Video Technology},
  32(2):634--645, 2021.

\bibitem[\protect\citeauthoryear{Shimizu \bgroup \em et al.\egroup
  }{2020}]{shimizu2020hyperbolic}
Ryohei Shimizu, Yusuke Mukuta, and Tatsuya Harada.
\newblock Hyperbolic neural networks++.
\newblock {\em arXiv preprint arXiv:2006.08210}, 2020.

\bibitem[\protect\citeauthoryear{Smith \bgroup \em et al.\egroup
  }{2017}]{smith2017federated}
Virginia Smith, Chao-Kai Chiang, Maziar Sanjabi, and Ameet~S Talwalkar.
\newblock Federated multi-task learning.
\newblock {\em NeurIPS}, 30, 2017.

\bibitem[\protect\citeauthoryear{T~Dinh \bgroup \em et al.\egroup
  }{2020}]{t2020personalized}
Canh T~Dinh, Nguyen Tran, and Josh Nguyen.
\newblock Personalized federated learning with moreau envelopes.
\newblock {\em NeurIPS}, 33:21394--21405, 2020.

\bibitem[\protect\citeauthoryear{Tammes}{1930}]{tammes1930origin}
Pieter Merkus~Lambertus Tammes.
\newblock On the origin of number and arrangement of the places of exit on the
  surface of pollen-grains.
\newblock {\em Recueil des travaux botaniques n{\'e}erlandais}, 27(1):1--84,
  1930.

\bibitem[\protect\citeauthoryear{Tan \bgroup \em et al.\egroup
  }{2022a}]{tan2022towards}
Alysa~Ziying Tan, Han Yu, Lizhen Cui, and Qiang Yang.
\newblock Towards personalized federated learning.
\newblock {\em IEEE Transactions on Neural Networks and Learning Systems},
  2022.

\bibitem[\protect\citeauthoryear{Tan \bgroup \em et al.\egroup
  }{2022b}]{tan2022enhancing}
Yanchao Tan, Carl Yang, Xiangyu Wei, Chaochao Chen, Longfei Li, and Xiaolin
  Zheng.
\newblock Enhancing recommendation with automated tag taxonomy construction in
  hyperbolic space.
\newblock In {\em ICDE}, pages 1180--1192. IEEE, 2022.

\bibitem[\protect\citeauthoryear{Weber \bgroup \em et al.\egroup
  }{2020}]{weber2020robust}
Melanie Weber, Manzil Zaheer, Ankit~Singh Rawat, Aditya~K Menon, and Sanjiv
  Kumar.
\newblock Robust large-margin learning in hyperbolic space.
\newblock {\em NeurIPS}, 33:17863--17873, 2020.

\bibitem[\protect\citeauthoryear{Xiao \bgroup \em et al.\egroup
  }{2017}]{xiao2017fashion}
Han Xiao, Kashif Rasul, and Roland Vollgraf.
\newblock Fashion-mnist: a novel image dataset for benchmarking machine
  learning algorithms.
\newblock {\em arXiv preprint arXiv:1708.07747}, 2017.

\bibitem[\protect\citeauthoryear{Zhao \bgroup \em et al.\egroup
  }{2018}]{zhao2018federated}
Yue Zhao, Meng Li, Liangzhen Lai, Naveen Suda, Damon Civin, and Vikas Chandra.
\newblock Federated learning with non-iid data.
\newblock {\em arXiv preprint arXiv:1806.00582}, 2018.

\end{thebibliography}

\appendix
\section{Datasets}
\begin{figure*}[t]
\centering
\subfigure[EMNIST $\alpha=0.1$]{
\begin{minipage}{0.45\columnwidth}{
\centering
\label{fig:emnist_0.1}
\includegraphics[scale=0.3]{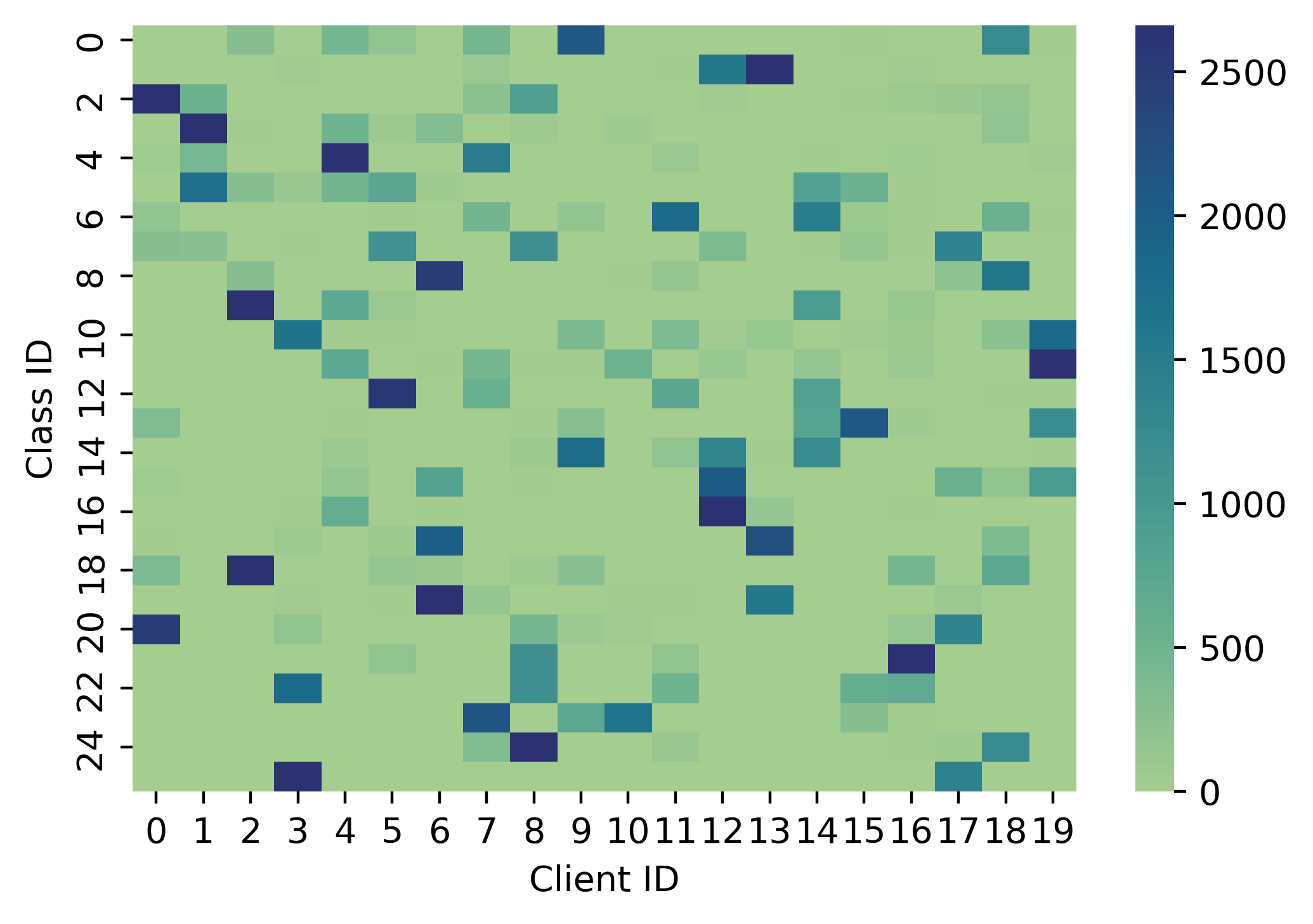} % 
}
\end{minipage}
}
\subfigure[FMNIST $\alpha=0.1$]{
\begin{minipage}{0.45\columnwidth}{
\centering
\label{fig:fmnist_0.1}
\includegraphics[scale=0.3]{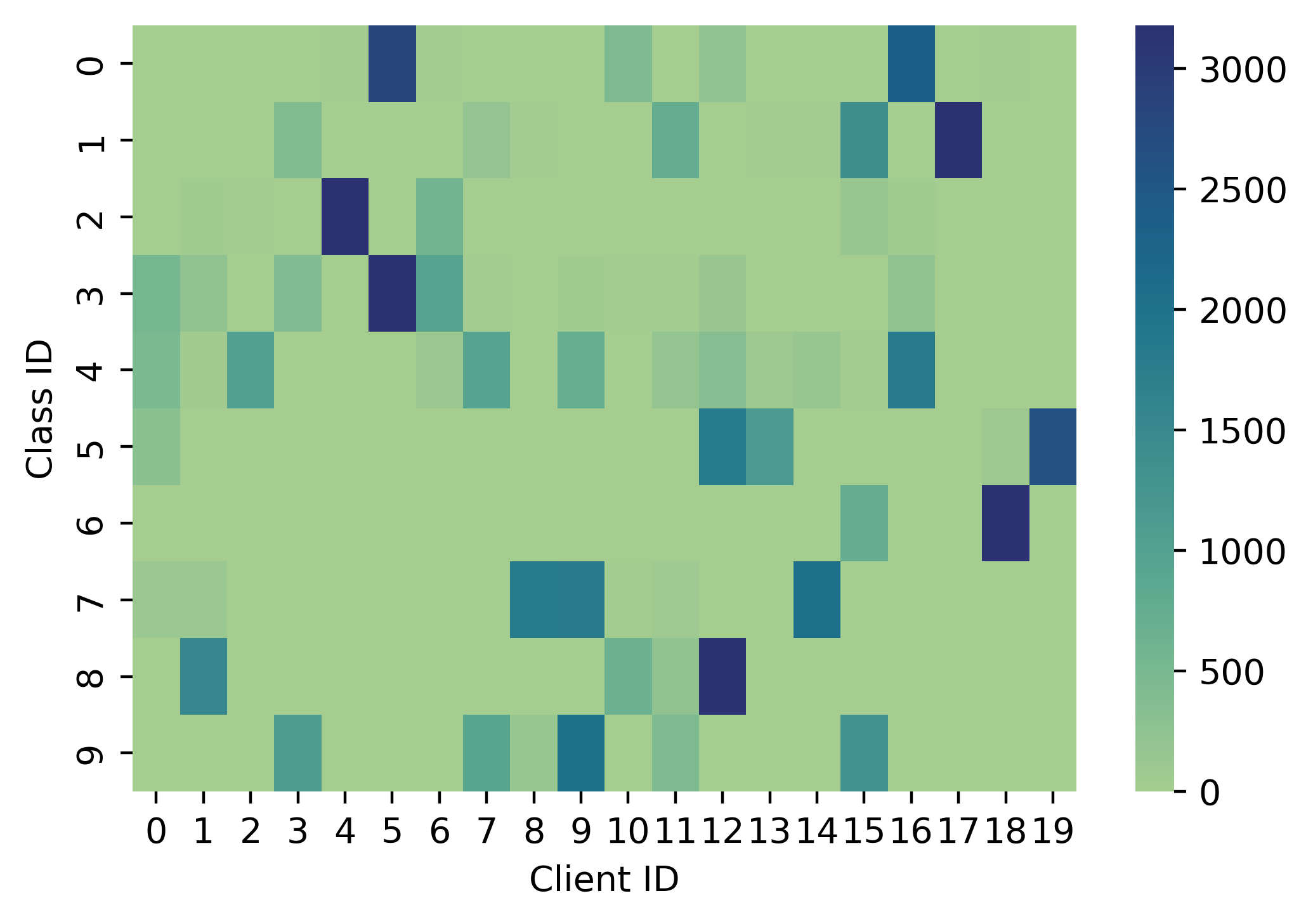} % 
}
\end{minipage}
}
\subfigure[Cifar10 $\alpha=0.1$]{
\begin{minipage}{0.45\columnwidth}{
\centering
\label{fig:cifar10_0.1}
\includegraphics[scale=0.3]{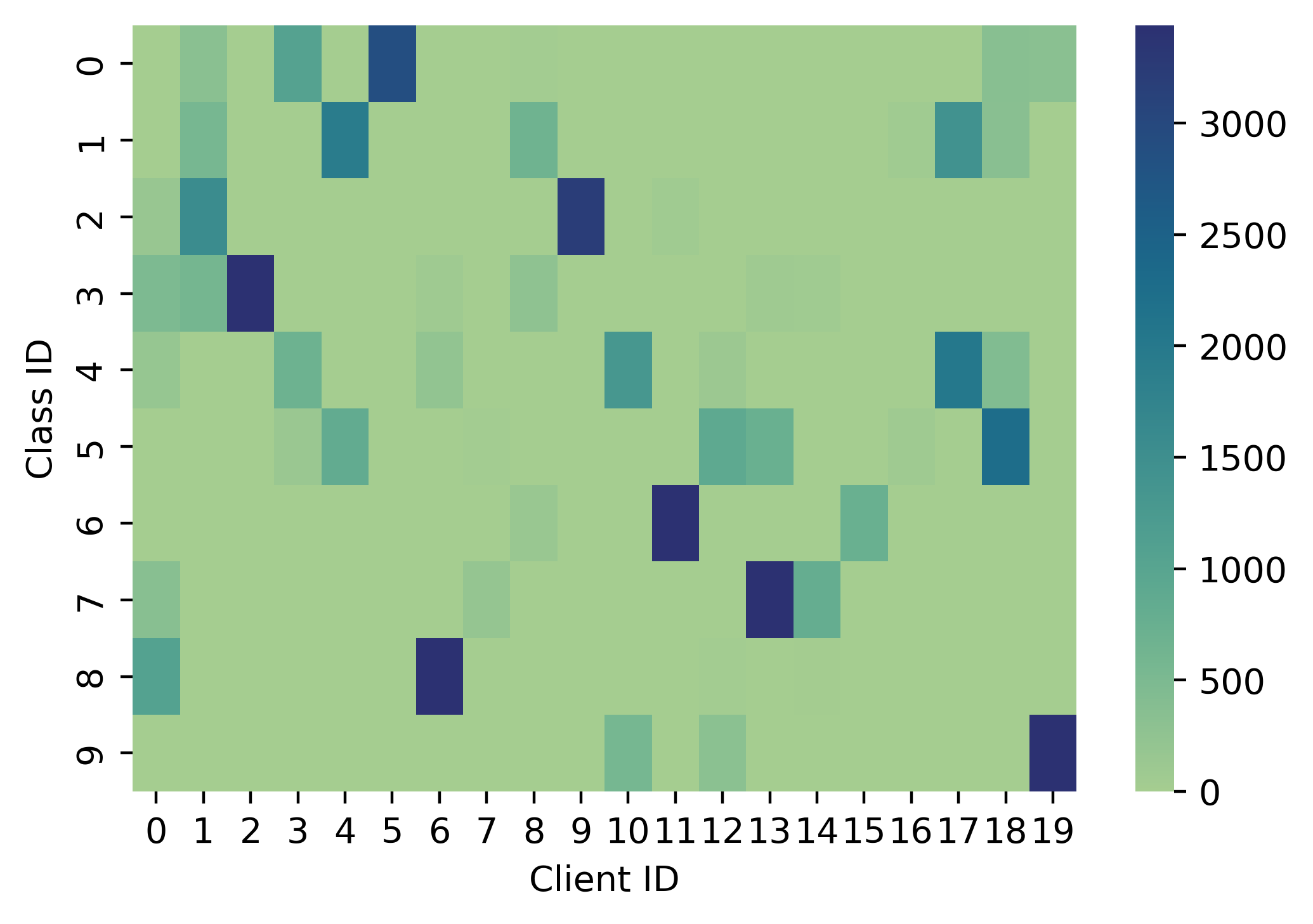} % 
}
\end{minipage}
}
\subfigure[Cifar100 $\alpha=0.1$]{
\begin{minipage}{0.45\columnwidth}{
\centering
\label{fig:cifar100_0.1}
\includegraphics[scale=0.3]{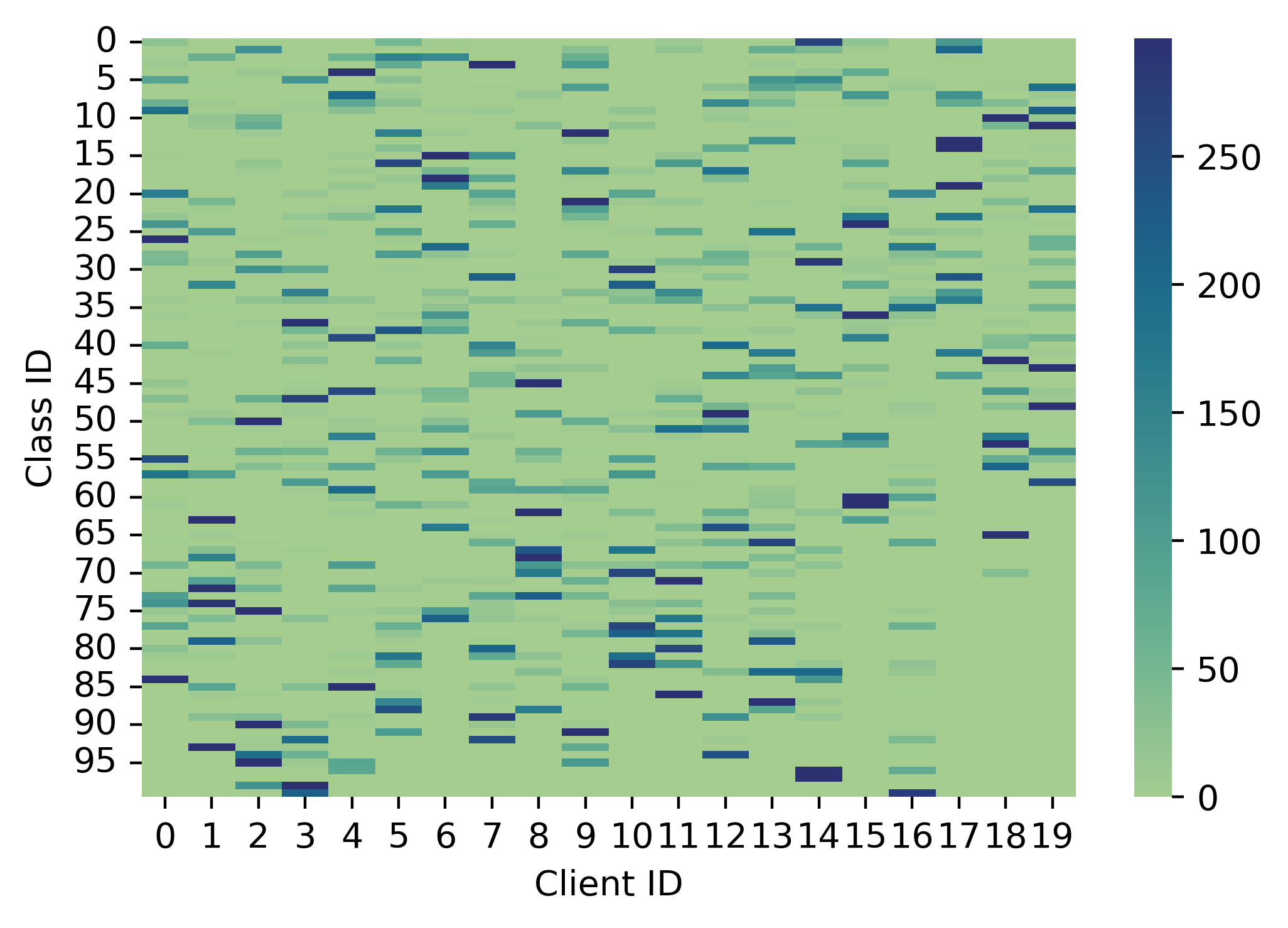} % 
}
\end{minipage}
}
\vspace{-0.1cm}
\caption{Distributions of $\alpha=0.1$.}
\label{fig:alpha01}
\end{figure*}

\begin{figure*}[t]
\centering
\subfigure[EMNIST $\alpha=0.5$]{
\begin{minipage}{0.45\columnwidth}{
\centering
\label{fig:emnist_0.5}
\includegraphics[scale=0.3]{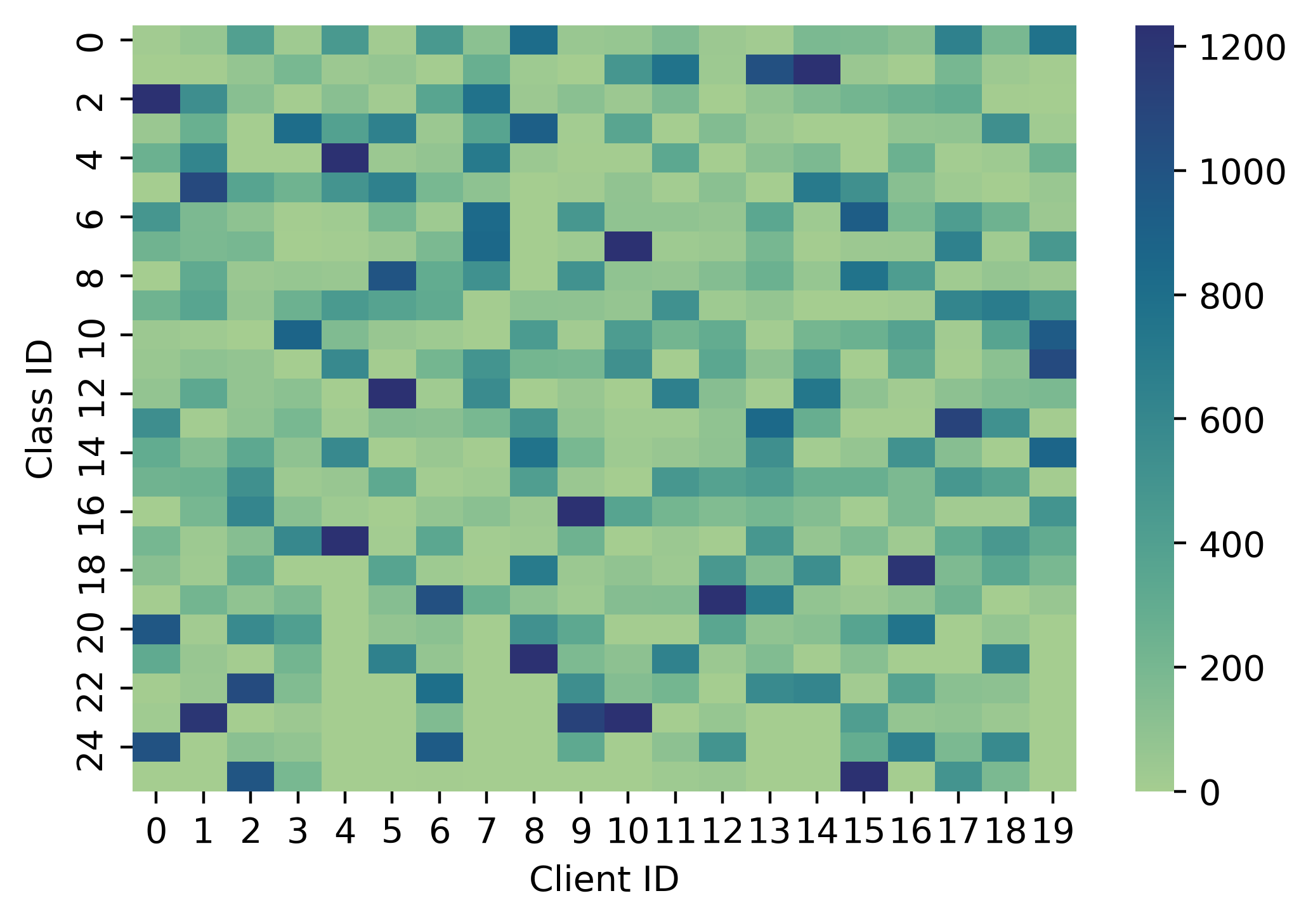} % 
}
\end{minipage}
}
\subfigure[FMNIST $\alpha=0.5$]{
\begin{minipage}{0.45\columnwidth}{
\centering
\label{fig:fmnist_0.5}
\includegraphics[scale=0.3]{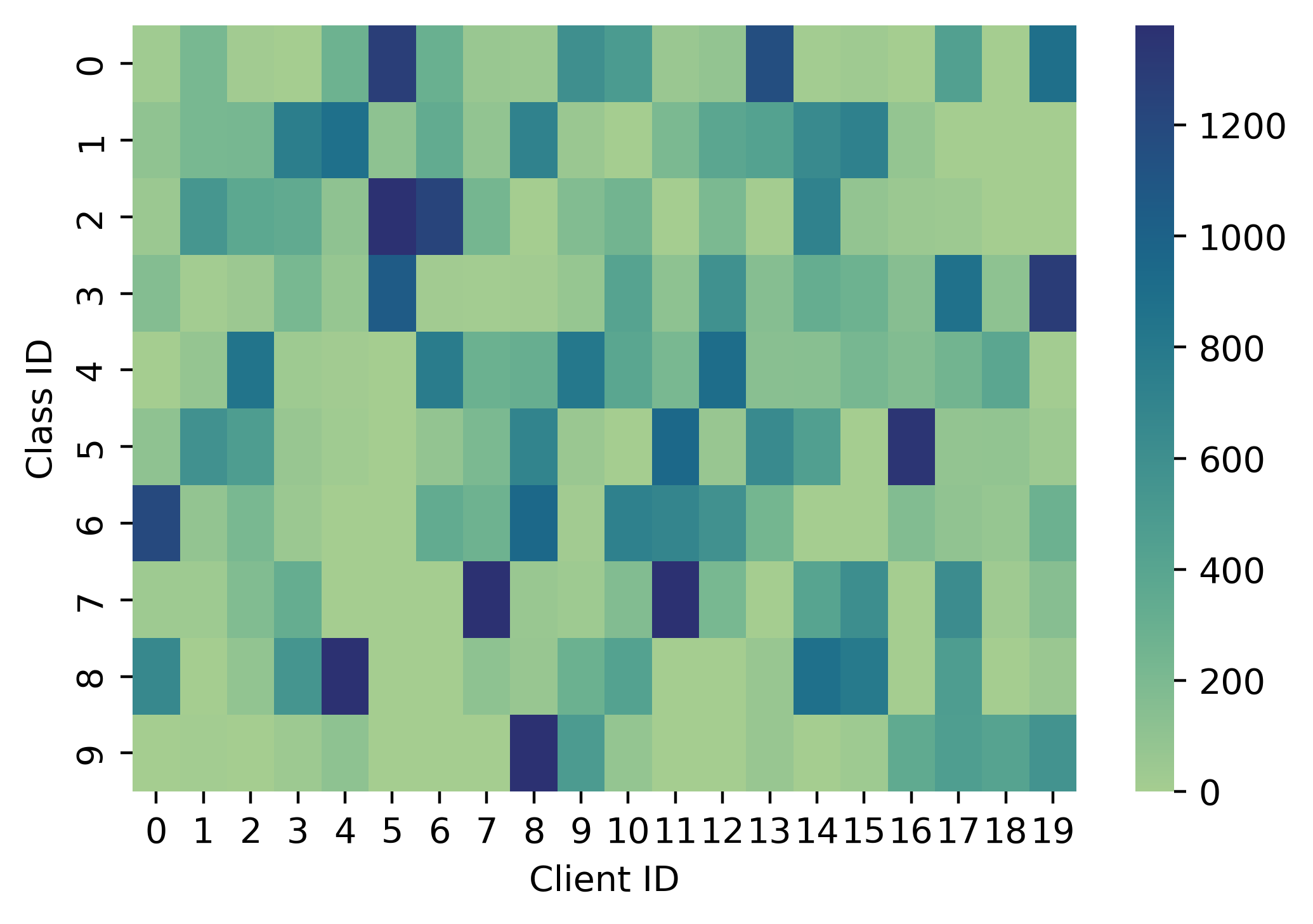} % 
}
\end{minipage}
}
\subfigure[Cifar10 $\alpha=0.5$]{
\begin{minipage}{0.45\columnwidth}{
\centering
\label{fig:cifar10_0.5}
\includegraphics[scale=0.3]{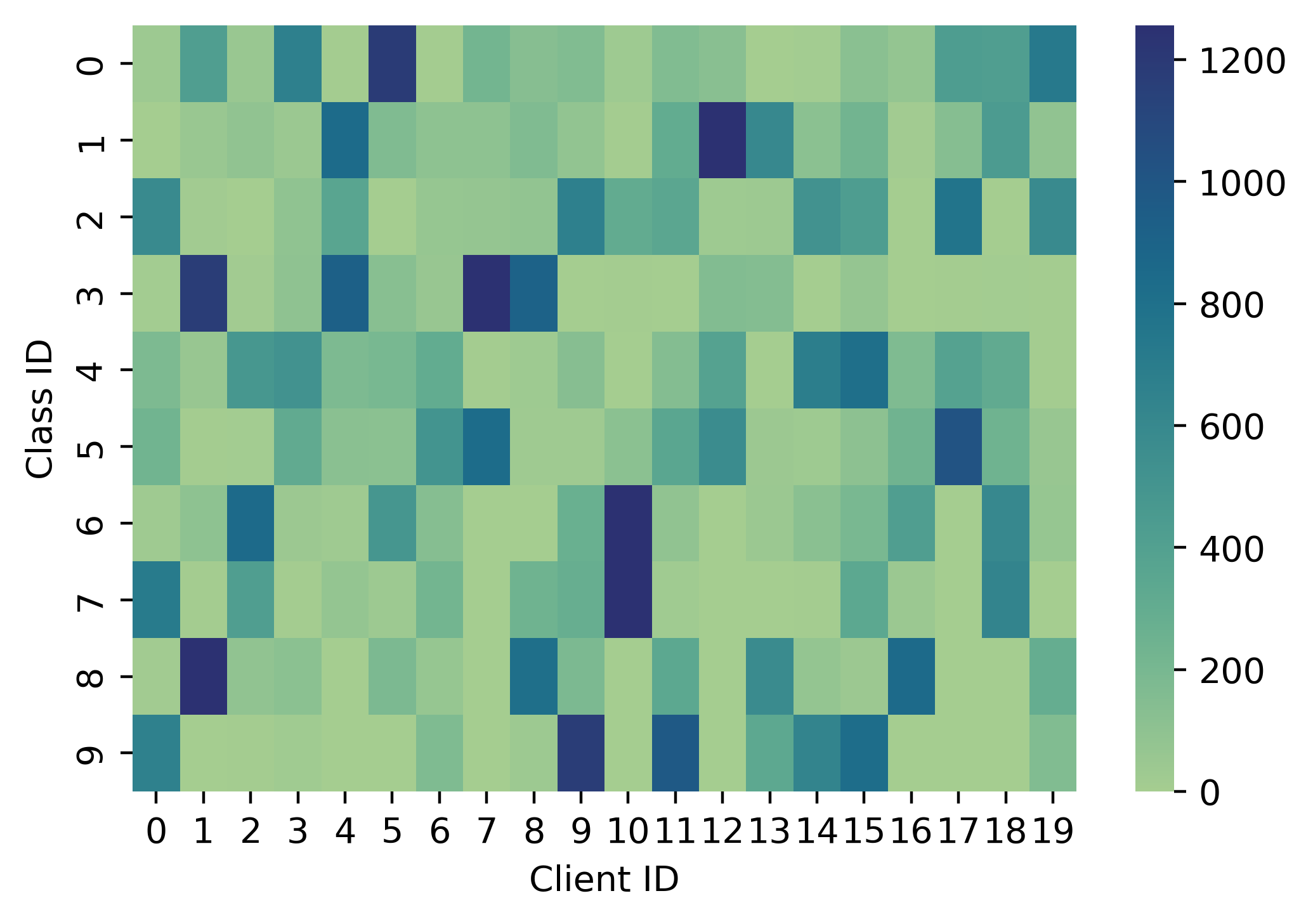} % 
}
\end{minipage}
}
\subfigure[Cifar100 $\alpha=0.1$]{
\begin{minipage}{0.45\columnwidth}{
\centering
\label{fig:cifar100_0.5}
\includegraphics[scale=0.3]{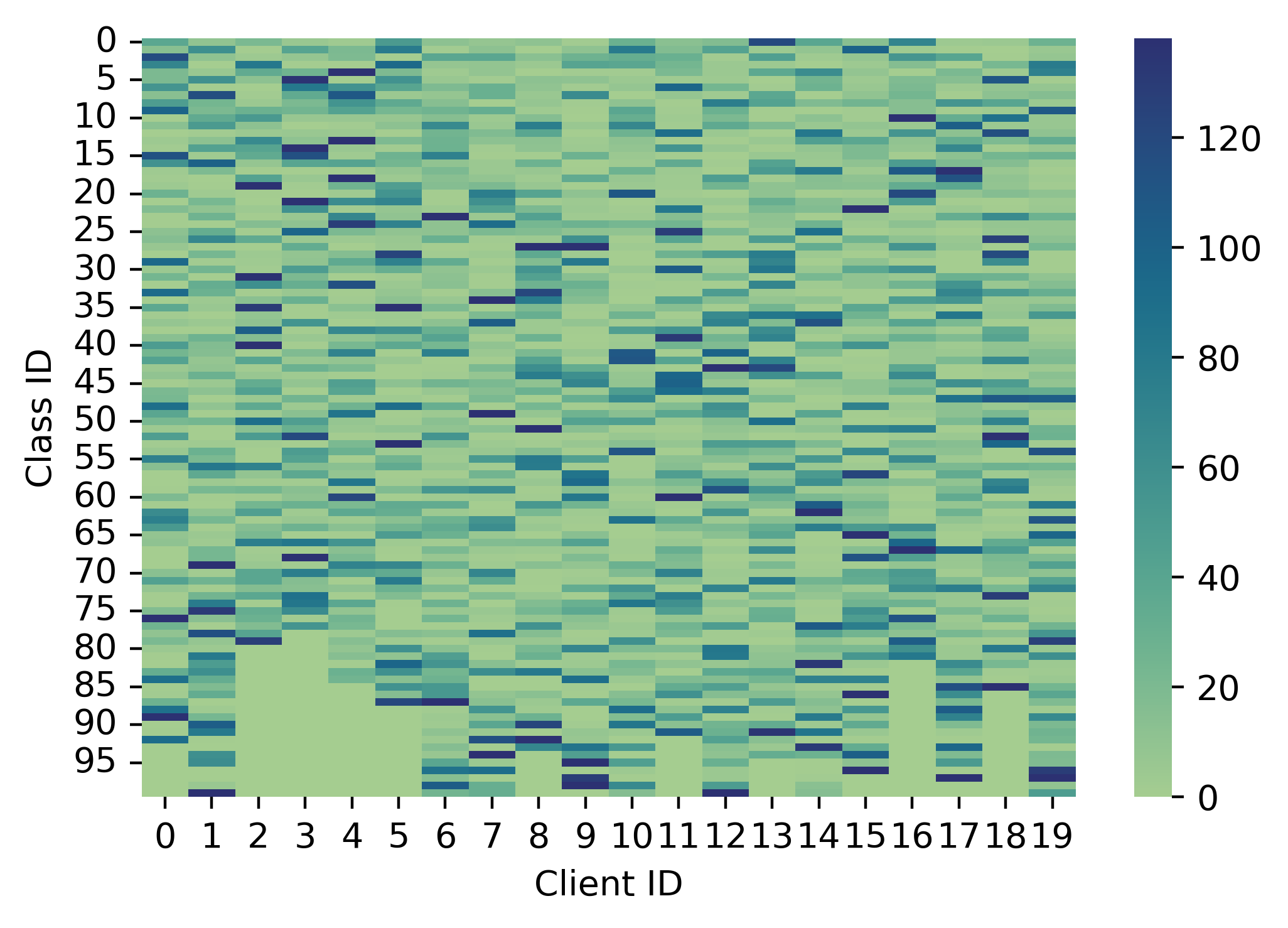} % 
}
\end{minipage}
}
\vspace{-0.1cm}
\caption{Distributions of $\alpha=0.5$.}
\label{fig:alpha05}
\end{figure*}

\begin{figure*}[h]
\centering
\subfigure[EMNIST $\alpha=5$]{
\begin{minipage}{0.45\columnwidth}{
\centering
\label{fig:emnist_5}
\includegraphics[scale=0.3]{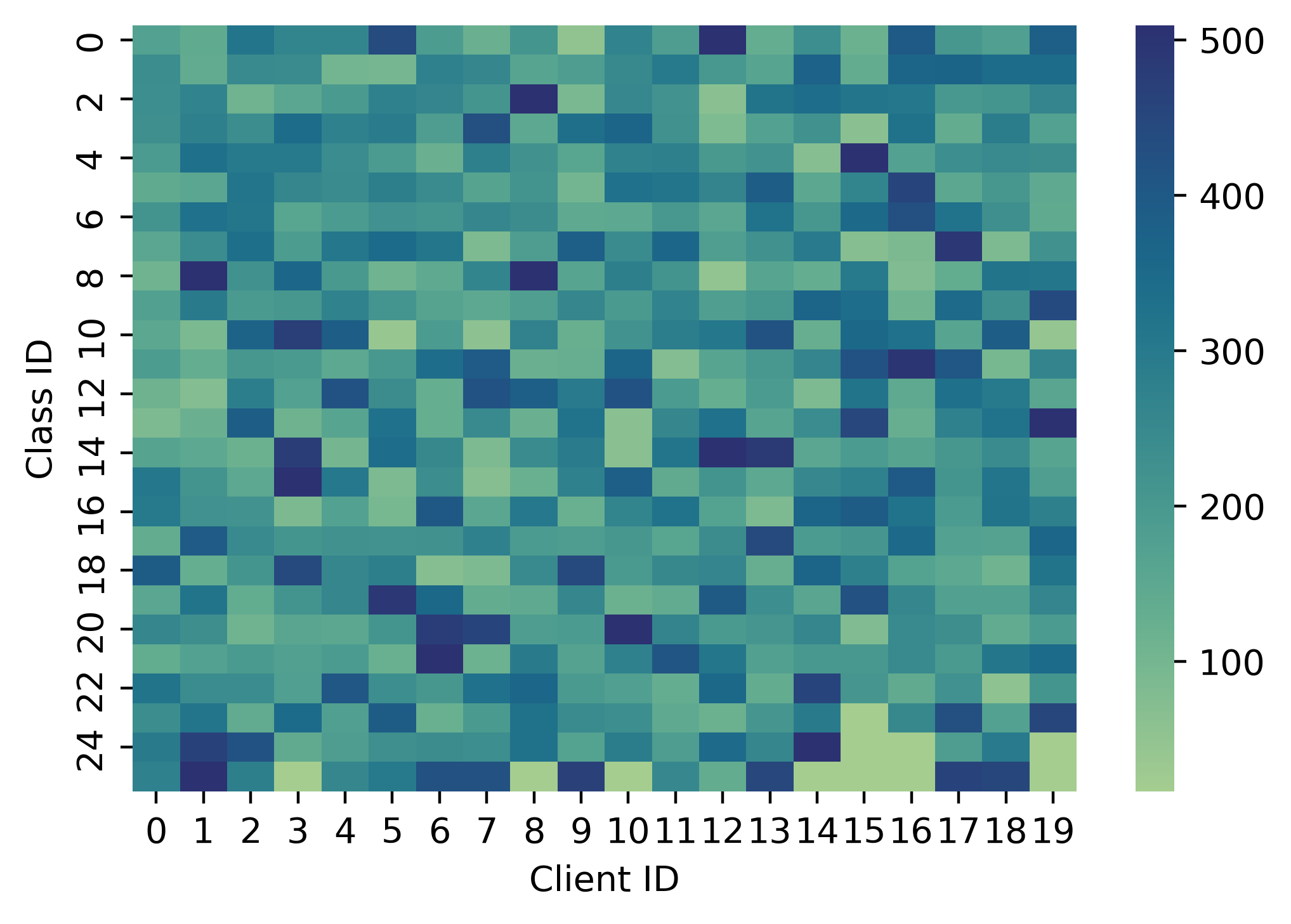} % 
}
\end{minipage}
}
\subfigure[FMNIST $\alpha=5$]{
\begin{minipage}{0.45\columnwidth}{
\centering
\label{fig:fmnist_5}
\includegraphics[scale=0.3]{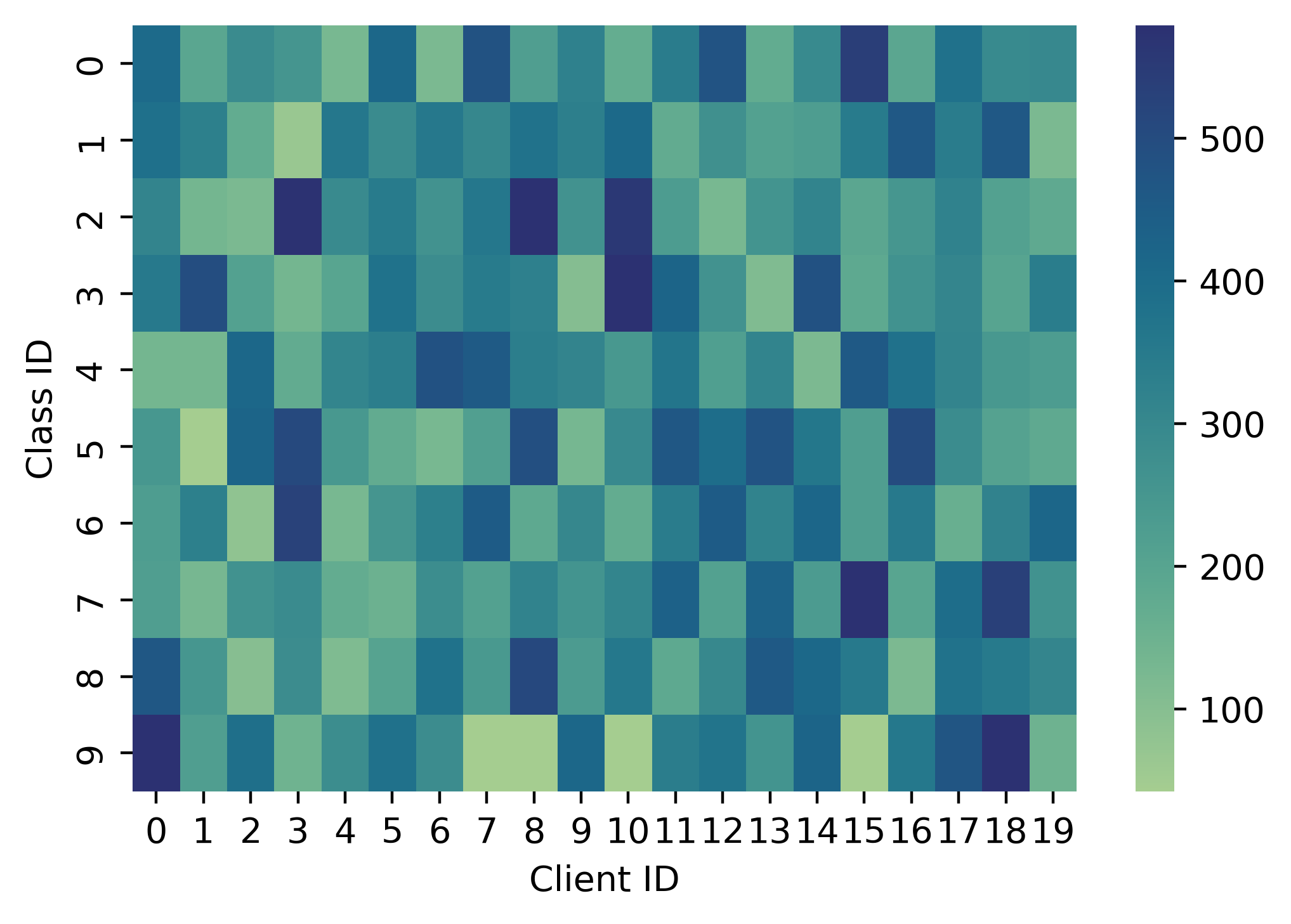} % 
}
\end{minipage}
}
\subfigure[Cifar10 $\alpha=5$]{
\begin{minipage}{0.45\columnwidth}{
\centering
\label{fig:cifar10_5}
\includegraphics[scale=0.3]{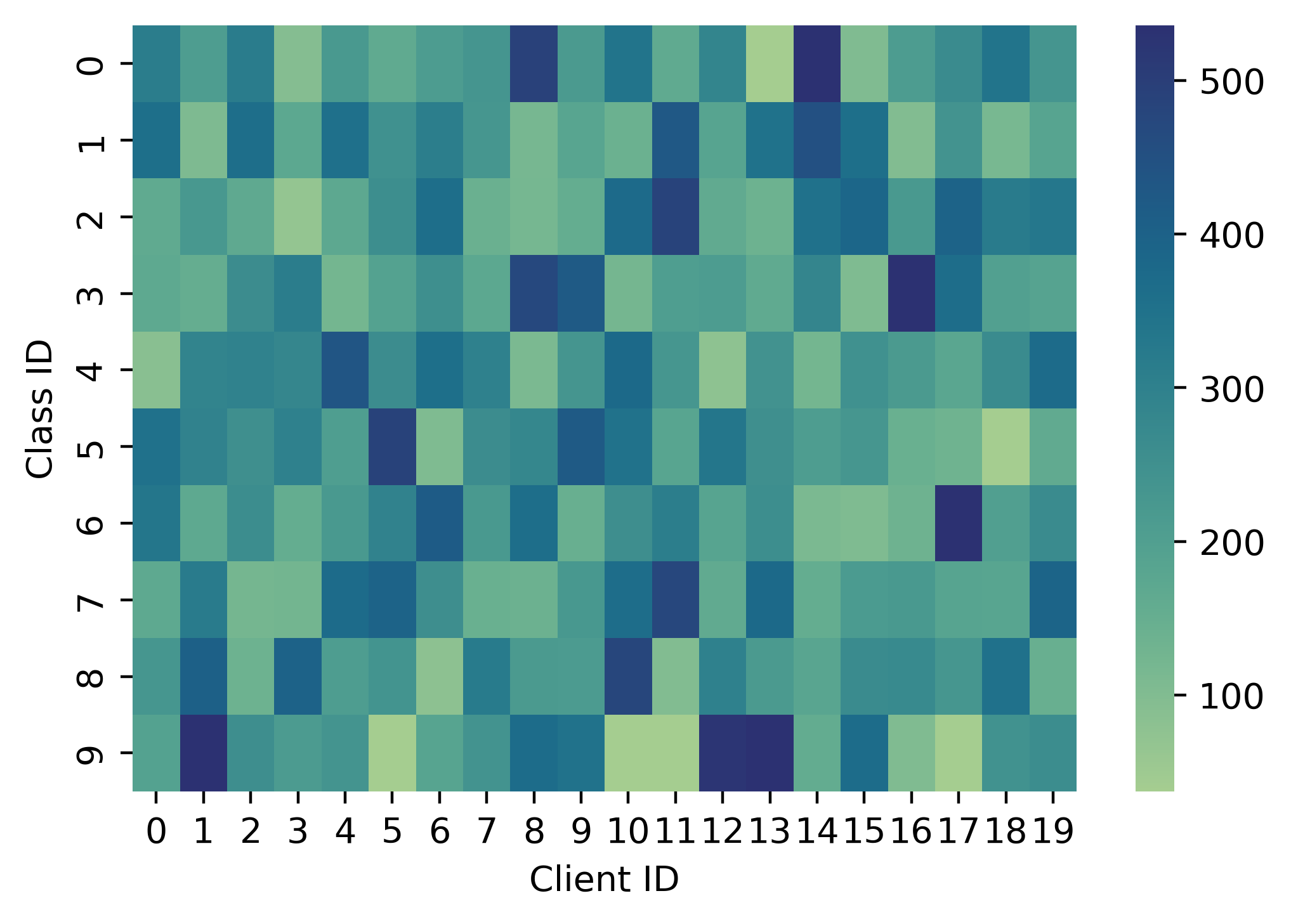} % 
}
\end{minipage}
}
\subfigure[Cifar100 $\alpha=5$]{
\begin{minipage}{0.45\columnwidth}{
\centering
\label{fig:cifar100_5}
\includegraphics[scale=0.3]{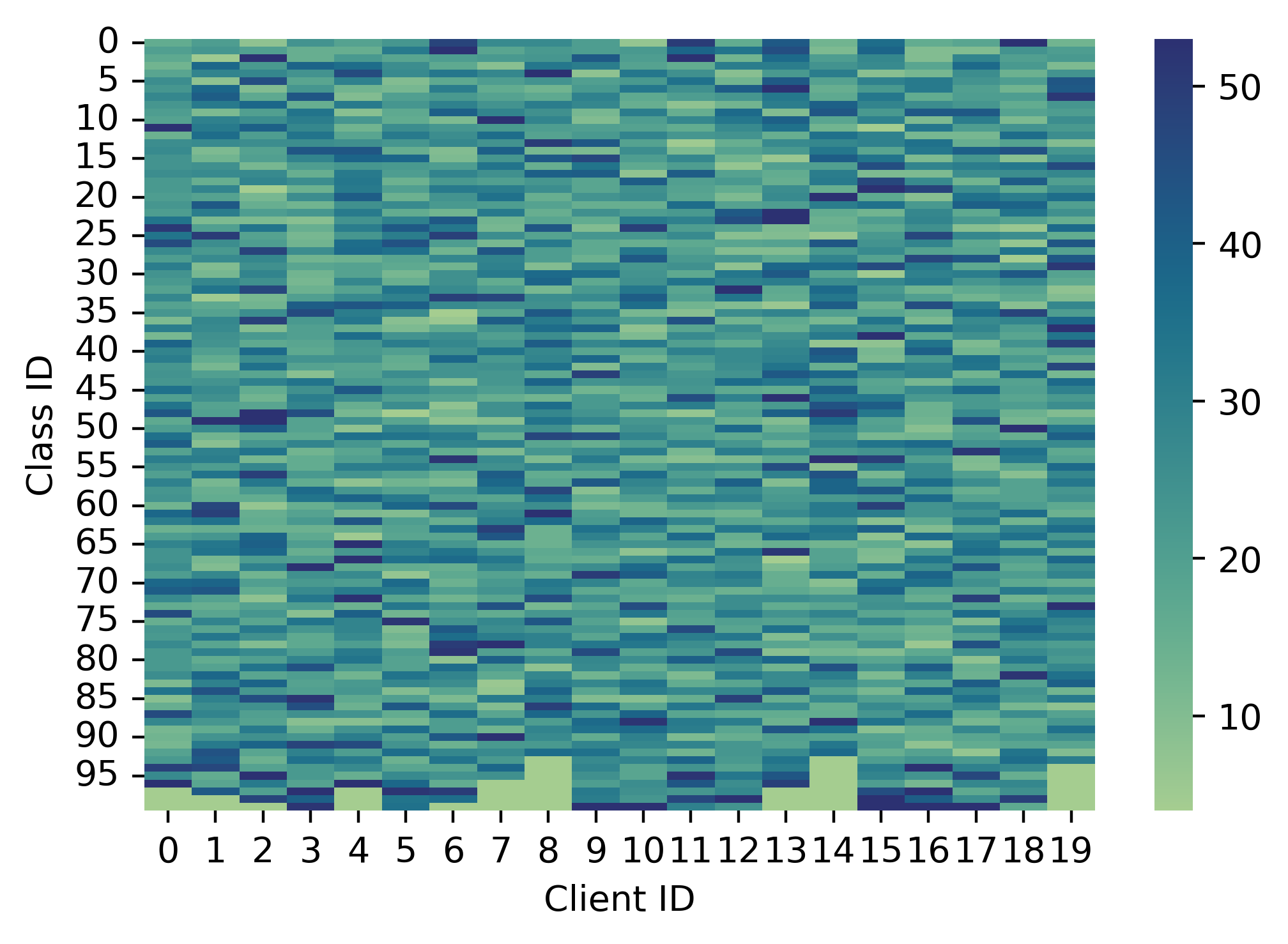} % 
}
\end{minipage}
}
\vspace{-0.1cm}
\caption{Distributions of $\alpha=5$.}
\label{fig:alpha5}
\end{figure*}

We visualize the data distributions under different degrees of non-IID, i.e., $\alpha =\{0.1, 0.5, 5\}$ for EMNIST, FMNIST, Cifar10, and Cifar100 in Fig.~\ref{fig:alpha01}-Fig.~\ref{fig:alpha5}.

\section{Details of Comparison Methods}
We compare \modelname~with three categories of state-of-the-art approaches according to their optimization goals, i.e., (1) optimizing global model: \textbf{FedAvg}~\cite{mcmahan2017communication}, \textbf{FedProx}~\cite{li2020federated}, \textbf{SCAFFOLD}~\cite{karimireddy2019scaffold}, \textbf{FedDYN}~\cite{acar2020federated}, \textbf{MOON}~\cite{li2021model}, (2) optimizing local personalized models: \textbf{FedMTL}~\cite{smith2017federated}, \textbf{FedPer}~\cite{arivazhagan2019federated}, \textbf{pFedMe}~\cite{t2020personalized}, \textbf{Ditto}~\cite{li2021ditto},
\textbf{APPLE}~\cite{luo2021adapt}, and (3) optimizing both global and local models: \textbf{Fed-RoD}~\cite{chen2021bridging}, \textbf{FedBABU}~\cite{oh2021fedbabu}, \textbf{SphereFed}~\cite{dong2022spherefed}. 
\textbf{FedAvg} is the first vanilla federated learning framework to collaborate among server and clients.
\textbf{FedProx} takes a proximal term to regularize the change from global model to the local model.
\textbf{SCAFFOLD} considers the variance of global model and local model when updating local gradients.
\textbf{FedDYN} applies an dynamic regularizer to pull local model close to the global model, while to push local model away from  historical local model.
\textbf{MOON} introduces contrastive learning to federated learning.
\textbf{FedMTL} is an algorithm that takes personalized learning as a multi-task learning objective.
\textbf{FedPer} decouples the global representation and local personalized classification.
\textbf{pFedMe} uses Moreau envelopes as clients’ regularized loss functions to decouple personalized model optimization from the global model learning. 
\textbf{Ditto} is a multi-task learning objective for federated learning that provides personalization while retaining similar efficiency. 
\textbf{APPLE} adaptively learns to personalize the client models.
\textbf{Fed-RoD}  explicitly decouples a model's dual duties with two prediction tasks. 
\textbf{FedBABU} only updates the representation body of the model during federated training, and the head is fine-tuned for personalization during the evaluation process. 
\textbf{SphereFed} is a hyperspherical federated learning framework to 
 address PFL problem.

We conduct all of the experiments on one NVIDIA RTX 3090 GPU with 24Gb Memory.
In the first category of methods, a hyper-parameter $\mu$ is mainly set to balance the local objective and regularization term (between the global model and local model).
According to the optimal results, we set $\mu =0.01,1,0.1, $ for FedProx,  MOON, and FedDYN, respectively.
Besides, we set $\tau=0.5$, the default temperature of MOON as specified in the original paper. 
For all of the second category of methods as well as the third category of methods, we follow the personalized way as their original setting.
We similarly finetune \modelname~and FedBABU with additional 5 local steps in P-FL.
For pFedMe, we set regularization parameter $\mu=1$ for controlling the strength of personalized model, computation complexity $k=5$, the interpolation parameter $\beta=1$ for global weights.
We choose hyper-parameter controlling the interpolation between global and local model , i.e., $\lambda=1 $ and local personalized iteration as 1 for Ditto.
In APPLE, we take the hyper-parameter $\mu=0.001$, and the round number  $L=0.2$ for loss scheduler.

\section{Theoretical Analysis}
\subsection{Convex Optimization}
\begin{figure*}[t]
\centering
\includegraphics[width=0.9\linewidth]{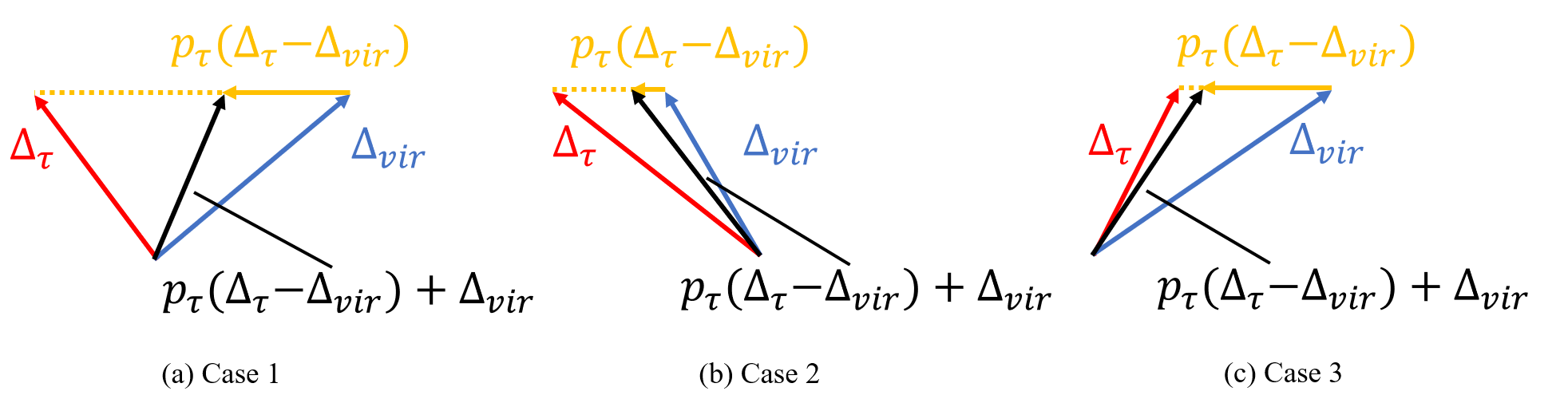}
\caption{Computational geometry of the deviations of the toughest client and the virtual client.}
\label{fig:CU}
\end{figure*}
% \begin{figure*}[t]
% \centering
% \subfigure[Case 1]{
% \begin{minipage}{0.6\columnwidth}{
% \centering
% \label{fig:case1}
% \includegraphics[scale=0.3]{figs/CU.png} % 
% }
% \end{minipage}
% }
% \subfigure[Case 2]{
% \begin{minipage}{0.6\columnwidth}{
% \centering
% \label{fig:case2}
% \includegraphics[scale=0.3]{figs/CU1.png} % 
% }
% \end{minipage}
% }
% \subfigure[Case 3]{
% \begin{minipage}{0.6\columnwidth}{
% \centering
% \label{fig:case3}
% \includegraphics[scale=0.3]{figs/CU2.png} % 
% }
% \end{minipage}
% }
% \vspace{-0.1cm}
% \caption{Distributions of $\alpha=0.5$.}
% \label{fig:alpha05}
% \end{figure*}

For the simplified optimization of the convex combination between the toughest deviation and the virtual deviation:
\begin{equation}
\operatornamewithlimits{min}\limits_{p_{\tau}\in [0,1]} \frac{1}{2} \|p_{\tau}\Delta_{{\tau}}+(1-p_{\tau})\Delta_{{\text{vir}}}\|^2_2,
\end{equation}
we rewrite it as:
\begin{equation}
\min _{p_{\tau} \in[0,1]} J = \frac{1}{2} \left\|p_{\tau}\left(\Delta_{{\tau}}- \Delta_{{\text{vir}}}\right)+\Delta_{{\text{vir}}}\right\|_2^2.
\end{equation}

Next, we can analyze the solution according to its geometry, i.e., analyzing the directions among the toughest deviation, the virtual deviation, and their difference.  
As the geometry suggests in Fig.~\ref{fig:CU}, the solution of finding the minimized norm is either a perpendicular vector or an edge case.
Thus we obtain the three cases as bellow:

\paragraph{Case 1} When it has the condition as:
\begin{equation}
\left\{ 
\begin{array}{l}  
\left(\Delta_{{\tau}}-\Delta_{{\text{vir}}}\right)^{\top} \cdot \Delta_{{\tau}}\ >0  \\
\left(\Delta_{{\tau}}-\Delta_{{\text{vir}}}\right)^{\top} \cdot \Delta_{{\text{vir}}} <0 ,
\end{array}
\right.
\end{equation}
Then we should take the differentiation $\frac{\partial J}{\partial p_{\tau}}=0$ and get:
\begin{equation}
\left(\Delta_{{\tau}}-\Delta_{{\text{vir}}}\right)^{\top}\left(p_{\tau} \Delta_{{\tau}} + (1-p_{\tau}) \Delta_{{\text{vir}}}\right)=0.
\end{equation}
Hence the solution is given:
\begin{equation}
\begin{gathered}
    p_{\tau}^*=\frac{\left(\Delta_{{\text{vir}}} - \Delta_{{\tau}} \right)^{\top} \cdot \Delta_{{\text{vir}}}}{\left\|\Delta_{{\tau}}- \Delta_{{\text{vir}}}\right\|_2^2} \quad \Leftrightarrow \quad \\
\Delta_{\boldsymbol{\theta}}^{t+1} =(1-p_{\tau}^*)\Delta_{{\text{vir}}}+ {p_{\tau}^*}\Delta_{\tau}.
\end{gathered}
\end{equation}
In this scenario, we find the perpendicular vector of the difference between the virtual deviation and the toughest deviation to obtain the minimum-norm.

\paragraph{Case 2} When it has the condition as:
\begin{equation}
\left(\Delta_{{\tau}}-\Delta_{{\text{vir}}}\right)^{\top} \cdot \Delta_{{\text{vir}}} \geq 0,
\end{equation}
the solution
\begin{equation}
p_{\tau}^*=0 \quad \Leftrightarrow \quad \Delta_{\boldsymbol{\theta}}^{t+1} =\Delta_{{\text{vir}}}.
\end{equation}
In this scenario, the $\left\|\Delta_{{\text{vir}}}\right\|_2^2$ is smaller than $\left\|\Delta_{{\tau}}\right\|_2^2$. The smaller deviation will hurdle the optimization on the virtual client. Therefore, we add weight on the virtual client.

\paragraph{Case 3} When it has the condition as:
\begin{equation}
\left(\Delta_{{\tau}}-\Delta_{{\text{vir}}}\right)^{\top} \cdot \Delta_{{\tau}} \leq 0,
\end{equation}
the solution
\begin{equation}
p_{\tau}^*=1, \quad \Leftrightarrow \quad \Delta_{\boldsymbol{\theta}}^{t+1} =\Delta_{{\tau}}.
\end{equation}
In this scenario, the $\left\|\Delta_{{\tau}}\right\|_2^2$ is smaller than $\left\|\Delta_{{\text{vir}}}\right\|_2^2$. The smaller deviation will hurdle the optimization on the toughest client. Therefore, we add weight on the toughest client.

\end{document}